\documentclass{article}
\usepackage{layout}

\usepackage{times}
\usepackage{soul}
\usepackage{url}
\usepackage[hidelinks]{hyperref}
\usepackage[utf8]{inputenc}
\usepackage[small]{caption}
\usepackage{graphicx}
\usepackage{amsmath}
\usepackage{amsthm}
\usepackage{booktabs}
\usepackage{algorithm}
\usepackage{algpseudocode}
\usepackage[switch]{lineno}
\usepackage{dblfloatfix}
\usepackage{amsfonts}
\usepackage{todonotes}
\usepackage[caption=false]{subfig} %

\title{CROP: Towards Distributional-Shift Robust Reinforcement Learning using Compact Reshaped Observation Processing
\footnote{Published at: Philipp Altmann, Fabian Ritz, Leonard Feuchtinger, Jonas Nüßlein, Claudia Linnhoff-Popien, and Thomy Phan. CROP: Towards Distributional-Shift Robust Reinforcement Learning Using Compact Reshaped Observation Processing. In \textit{Proceedings of the Thirty-Second International Joint Conference on Artificial Intelligence, IJCAI-23}, 8 2023, pages 3414–3422. DOI: \href{https://doi.org/10.24963/ijcai.2023/380}{10.24963/ijcai.2023/380}}

}

\author{
Philipp Altmann \and Leonard Feuchtinger \and Fabian Ritz \and Jonas Nüßlein \and \\ Thomy Phan \And Claudia Linnhoff-Popien
\affiliations LMU Munich 
\emails philipp.altmann@ifi.lmu.de
}

\author{
Philipp Altmann \and Fabian Ritz \and Leonard Feuchtinger \and Jonas Nüßlein \and \\ Claudia Linnhoff-Popien \And Thomy Phan
\affiliations LMU Munich 
\emails philipp.altmann@ifi.lmu.de
}

\begin{document}

\maketitle

\begin{abstract}
The safe application of reinforcement learning (RL) requires generalization from limited training data to unseen scenarios. 
Yet, fulfilling tasks under changing circumstances is a key challenge in RL. 
Current state-of-the-art approaches for generalization apply data augmentation techniques to increase the diversity of training data. 
Even though this prevents overfitting to the training environment(s), it hinders policy optimization.
Crafting a suitable observation, only containing crucial information, has been shown to be a challenging task itself.
To improve data efficiency and generalization capabilities, we propose Compact Reshaped Observation Processing (CROP) to reduce the state information used for policy optimization.
By providing only relevant information, overfitting to a specific training layout is precluded and generalization to unseen environments is improved.
We formulate three CROPs that can be applied to fully observable observation- and action-spaces and provide methodical foundation. 
We empirically show the improvements of CROP in a distributionally shifted safety gridworld.
We furthermore provide benchmark comparisons to full observability and data-augmentation in two different-sized procedurally generated mazes. 
\end{abstract}

\section{Introduction}
To safely deploy \textit{machine learning} (ML) methods in real-world scenarios, generalization is an important challenge.
As training data cannot contain all possible situations in general, ML methods should be able to generalize to unseen samples instead of overfitting to the training data. 
More specifically, their learned behavior should be robust to scenarios not included in the training data, often also refereed to as out-of-distribution (OOD) generalization \cite{hendrycks2021many,quinonero2008dataset}.
Whilst important for supervised- and unsupervised-learning tasks, said generalization is especially important for the safe application of \textit{reinforcement learning} (RL), where unexpected observations may cause unintended and potentially unsafe behavior \cite{amodei2016concrete}.
For instance, a shift might be present when transferring a robotics model from the training simulation to the real-world \cite{zhao2020sim}. 
Generally, increased generalization from a few examples, also referred to as frew-shot learning, is induced by altering the training data to increase its diversity and impede overfitting \cite{wang2020generalizing}.
Those \textit{data augmentation} techniques have been successfully applied to both supervised and reinforcement learning by increasing the information used for training the model \cite{laskin2020reinforcement}.
However, while the enlargement of the data may prevent overfitting and thus improve robustness to OOD samples, the increased complexity also increases training difficulty and hinders optimal convergence. 
On the contrary, despite tracing insufficient generalization to overfitting to the training data, occurrences of benign overfitting have been shown and characterized \cite{bartlett2020benign}.
Also, the states observed by the agent might not resemble an optimal representation, an ``issues that are so often critical to successful applications'' of reinforcement learning \cite{sutton2018reinforcement}. 

Extending on these insights, we propose to apply \textit{Compact Reshaped Observation Processing} (CROP) to reduce the observation such that presumably irrelevant details are removed and the remaining information is sufficient for learning a robust policy that generalizes to unseen shifted observations. 
Overall, we provide the following contribution: 
\begin{itemize}
\item We formulate three concrete CROP methods applicable to fully observable state and action spaces, reducing the information with regards to the agent's position within the environment, the agent's action space and affect, and the objects within the environment (\autoref{fig:CROP}). 
\item We provide proof-of-concept and show the strengths of each CROP over the full observation in a safety gridworld training environment and a shifted test environment. 
\item We evaluate the proposed CROPs and compare their effect in zero-shot generalization to the state-of the art of data augmentation in randomly generated mazes. 
\end{itemize}

\section{Background}

\subsection{Distributional Shifts}\label{sec:ds}
One key assumption in ML is that the data for training, testing and runtime is independent and identically distributed \cite{bishop2006pattern}.
Then, good performance in individual stages of the process would imply a similar performance in all other stages \cite{malinin2021shifts}.
In practice, however, samples encountered during runtime may be out-of-distribution, e.g. due to sensors degrading over time \cite{hendrycks2021many}. 
\textit{Distributional Shifts} describe a problem where all samples are out-of-distribution, i.e. the whole distribution shifts during runtime \cite{quinonero2008dataset}. 
This shift might be a slight unnoticeable change, or a significant alteration. 
Therefore, the key assumption of identically distributed data does not hold and performance may be impacted negatively.
In safety-critical environments, this can cause severe issues, especially if the ML model makes unintended mistakes due to unexpected changes \cite{leike2017ai}.
This paper focuses on solutions referred to as \textit{zero-shot generalization}, where, given limited amount of training data, generally applicable solutions for conceptually similar problems shall be inferred.

\subsection{Problem Formulation}\label{sec:mdp}
Fully observable decision making problems can be formulated as \emph{Markov Decision Process (MDP)} $M_{\textit{MDP}} = \langle \mathcal{S}, \mathcal{A}, \mathcal{T}, \mathcal{R} \rangle$ , where $\mathcal{S}$ is a set of states $s_t$ from a feature space $\mathcal{F}$, $\mathcal{A}$ is a set of actions $a_t$, $\mathcal{T}$ is the transition function $\mathcal{T} : \mathcal{S} \times \mathcal{A} \mapsto \Delta(\mathcal{S})$ and $\mathcal{R}$ the reward function $\mathcal{R} : \mathcal{S} \times \mathcal{A} \mapsto \mathbb{R}$ \cite{puterman1990markov}.
Additionally, we consider a shifted set of states $\mathcal{S}^+\neq\mathcal{S}$ that may be encountered alternatively. 

The goal is to find a \emph{policy} $\pi : \mathcal{S} \mapsto\Delta(\mathcal{A})$, which maximizes the value $V^{\pi}(s_t) = \mathbb{E}_{\pi}[G_t|s_t]$ for all $s_t \in \mathcal{S}$, where $G_t = \sum_{t=0}^{\infty}\gamma^{t}\mathcal{R}(s_t, a_t)$ is the \emph{return} and $\gamma = [0,1]$ is the discount factor. An \emph{optimal policy} $\pi^{*}$ has the \emph{optimal value function} $V^{\pi^{*}} = V^{*}$ satisfying $V^{*} \geq V^{\pi'}$ for all $s_t \in \mathcal{S}$ and $\pi'$.

MDPs can be extended to \emph{Partially Observable Markov Decision Processes (POMDP)} $M_{\textit{POMDP}} = \langle \mathcal{S}, \mathcal{A}, \mathcal{T}, \mathcal{R}, \Omega, \mathcal{O}, b_{0} \rangle$, additionally consisting of a set $\Omega$ of observations $o_t$, observation function $\mathcal{O} : \mathcal{S} \times \mathcal{A} \mapsto \Delta(\Omega) $, and initial state distribution $b_0 \rightarrow \Delta(\mathcal{S})$, where the agent does not perceive the true state $s_t$ of the environment, but only a noisy observation $o_t \in \mathcal{O}$ according to $\mathcal{O}(o_t|s_t,a_{t-1})$ \cite{kaelbling1998planning}.

\subsection{Reinforcement Learning}\label{sec:rl}
\emph{Reinforcement learning (RL)} is an experience-based approach to find \emph{optimal policies} $\pi^{*}$ using experience tuples $e_t = \langle s_t, a_t, \mathcal{R}(s_t, a_t), s_{t+1} \rangle$.

\emph{Actor-critic} methods are common RL algorithms using a function approximator $\hat{\pi}_\theta \approx \pi^{*}$ with learnable parameters $\theta$, which are updated via gradient ascent according to gradient $g$ \cite{sutton2000policy}:
\begin{equation}\label{eq:actor_critic}
g = \hat{A}(s_{t},a_{t})\nabla_{\theta} \textit{log} \hat{\pi}_{\theta}(a_{t}|s_{t})
\end{equation}
where $\hat{A}(s_{t},a_{t}) = Q^{\hat{\pi}_{\theta}}(s_{t},a_{t}) - V^{\hat{\pi}_{\theta}}(s_{t})$ is the \emph{advantage} and $Q^{\hat{\pi}_{\theta}}$ is the \emph{action value function}. In practice, the return $G_t$ is used to approximate $Q^{\hat{\pi}_{\theta}}(s_{t},a_{t})$ \cite{mnih2016asynchronous}.

\emph{Proximal policy optimization (PPO)} is a modified version of the standard \textit{Advantage Actor-Critic} (A2C) method according to Eq. \ref{eq:actor_critic}, which iteratively minimizes a surrogate loss function $\mathcal{L}_{\textit{PPO}}$ to ensure stable learning \cite{schulman2017proximal}:
\begin{equation}\label{eq:ppo}
\mathcal{L}_{\textit{PPO}}(\theta) = \textit{min}(r_t \hat{A}(s_{t},a_{t}), \textit{clip}(r_t, 1 - \epsilon, 1 + \epsilon) \hat{A}(s_{t},a_{t}))
\end{equation}
where $r_t = \frac{\hat{\pi}_{\theta}(a_t|s_t)}{\hat{\pi}_{\theta, \textit{old}}(a_t|s_t)}$ is the importance sampling ratio of the current and old action probability with $\hat{\pi}_{\theta, \textit{old}}$ being the policy originally collecting experience samples $e_t$ for the update, and $\epsilon < 1$ is a clipping parameter, ensuring bounded changes $\theta$ to mitigate divergence.

\section{Related Work}
\subsection{Robustness and Generalization in RL}\label{sec:rob-gen}
Training robust RL agents that act reliably in unknown situations is a known challenge.
One branch of research deals with the recognition of distributional shift \cite{ramanan2021real} and unknown states \cite{pimentel2014review,thulasidasan2021effective}.
If distributional shift or unknown states are detected, one solution is to have the RL agent ask a (human) supervisor for advice \cite{najar2021reinforcement} or to adjust the RL agent, e.g., through further training.
In Adversarial RL \cite{pinto2017robust}, an opponent policy partially controls the agent during training, with the aim of minimizing the long-term reward.
The result is a policy that is more robust to changes in the environment, as it anticipates intervention.
Our approach does not detect distributional shift or include adversaries.
Instead, it aims to increase robustness via zero-shot generalization.

Another branch of research aims to improve generalization by avoiding overfitting to the training data.
Here, a variety of methods has been developed: stopping the training early \cite{raskutti2014early}, dropping random parts of the underlying neural network \cite{srivastava2014dropout}, or augmenting the training data with noise \cite{karystinos2000overfitting}.
Furthermore, training can be carried out in as many different environments as possible \cite{tobin2017domain,cobbe2020leveraging,gisslen2021adversarial}, such that an RL agent can not succeed by overfitting to a small number of trajectories and is forced to acquire transferable knowledge.
However, this requires a huge number of different environments, that are typically created by exhaustively generating variations of the same procedural environment.
This is not sample efficient and makes it difficult to create distinct tests.
However, the lack of distinct tests both diminishes the impact of distributional shift and the need for strong generalization to succeed.
Thus, we aim to improve generalization by training with a reduced set of more relevant training data.
\subsection{Data Augmentation in RL}\label{sec:rad}
Quantity and quality of the dataset heavily impact the training in ML \cite{ying2019overview}.
\textit{Data Augmentation} can be used to artificially increase the diversity in training data when only a limited amount of data is available.
The idea is to systematically modify the training data to avoid homogeneous structures \cite{shorten2019survey}.
In image recognition tasks, this is done by geometric transformations such as mirroring, rotating or hiding pixels.
This serves as a regularization against overfitting and increases data efficiency, ultimately improving generalization \cite{kostrikov2020image,laskin2020reinforcement}.

Various approaches adopted these ideas to RL \cite{raileanu2020automatic,chen2020gridmask,yarats2021mastering}.
Similar to our approach, \textit{Reinforcement Learning with Augmented Data} (RAD) \cite{laskin2020reinforcement} augments observations without domain specific knowledge or changes to the RL algorithms.
In addition to image based observations, RAD proposes two methods for state based observations: random amplitude scaling and Gaussian noise.
By creating more diverse training data, the authors report increased data efficiency and better generalization to unseen environments.

We propose three additional methods to augment state based observations that reduce the amount of distinct training observations instead of increasing it.
Improvements of the training are a welcome positive effect, but we focus on zero-shot generalization to unseen environments.
\subsection{Partial Observability and Invariants}\label{sec:po}
In real-world problems, RL agents often face incomplete and imperfect information \cite{choi2019deep} and thus may perceive different states as similar \cite{spaan2012partially}.
In such POMDPs, learning optimal policies with naive approaches is difficult, and the respective stochasticity is regarded as a fundamental challenge for RL \cite{vlassis2012computational,ghosh2021generalization,jaakkola1994reinforcement}.
However, we propose to train with limited state information on purpose, as this potentially mitigates the effect of directly perceiving the distributional shift and thus may improve robustness.
Naturally, the remaining state information must be sufficient to find an optimal policy.

Removing potentially irrelevant information to improve the training is common in other ML areas such as facial recognition \cite{chen2014cross}.
While less common in RL, recent approaches followed this concept and proposed to explicitly learn invariants.
\cite{zhang2020learning} showed that agents can learn observation representations in latent space which encode task-relevant information.
\cite{agarwal2021contrastive} use a policy similarity metric (states are similar if the optimal policy has a similar behaviour in that and future states) with a contrastive learning approach to learn policies invariant to observation variations.
\cite{mazoure2021cross} use clustering methods and self-supervised learning to define an auxiliary task, which is mapping behaviorally similar states to similar representations.
In fact, all these approaches require different observations from multiple training contexts and a complex nonlinear encoder that maps observations to a latent representation.
On the contrary, we rely on a less complex hand-crafted reduction of state information.

\begin{figure*}
  \centering
  \subfloat[Radius CROP]{
    \includegraphics[width=0.25\textwidth]{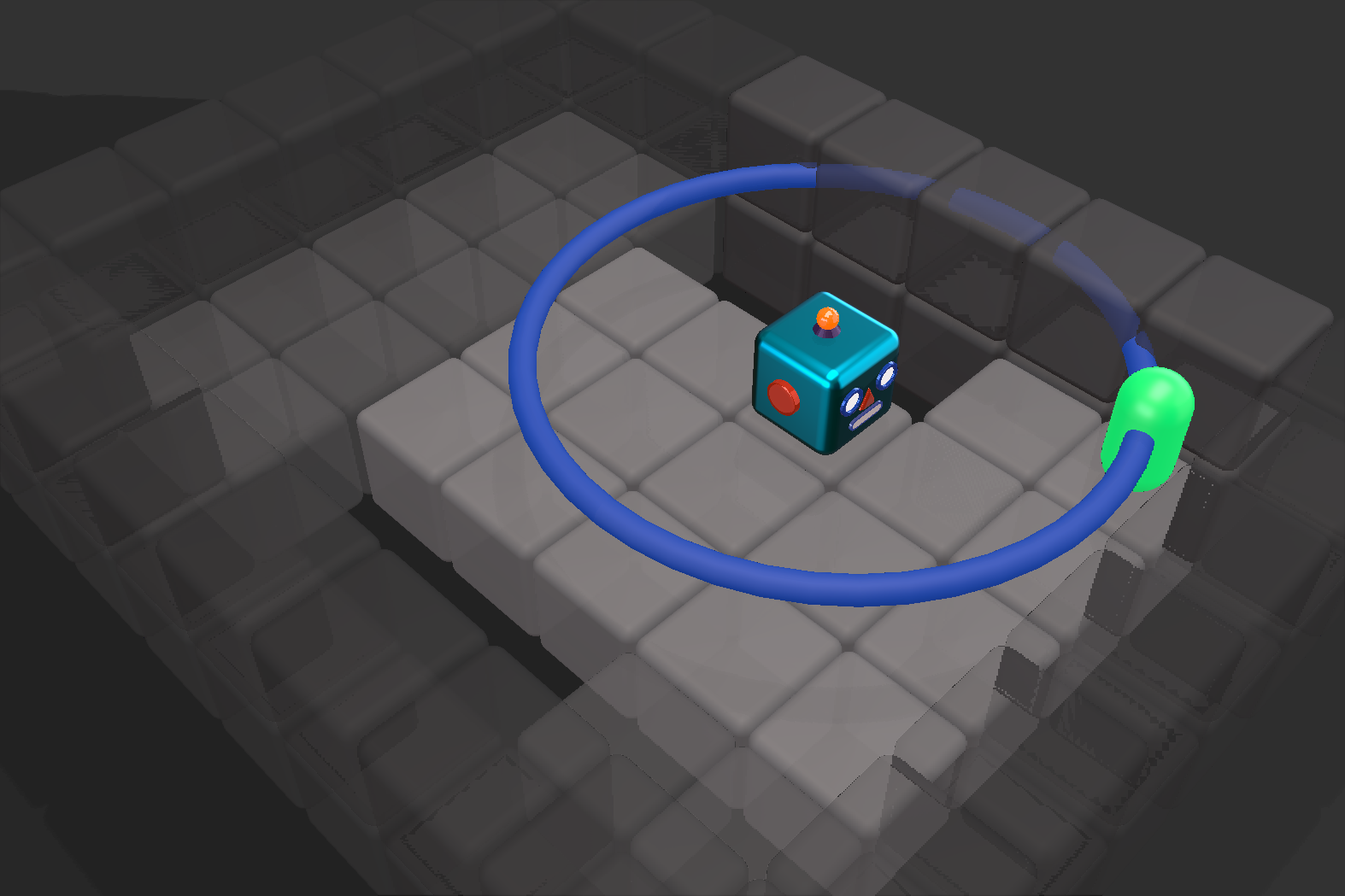}\label{fig:CROP:Radius}
  }
  \subfloat[Action CROP]{
    \includegraphics[width=0.25\textwidth]{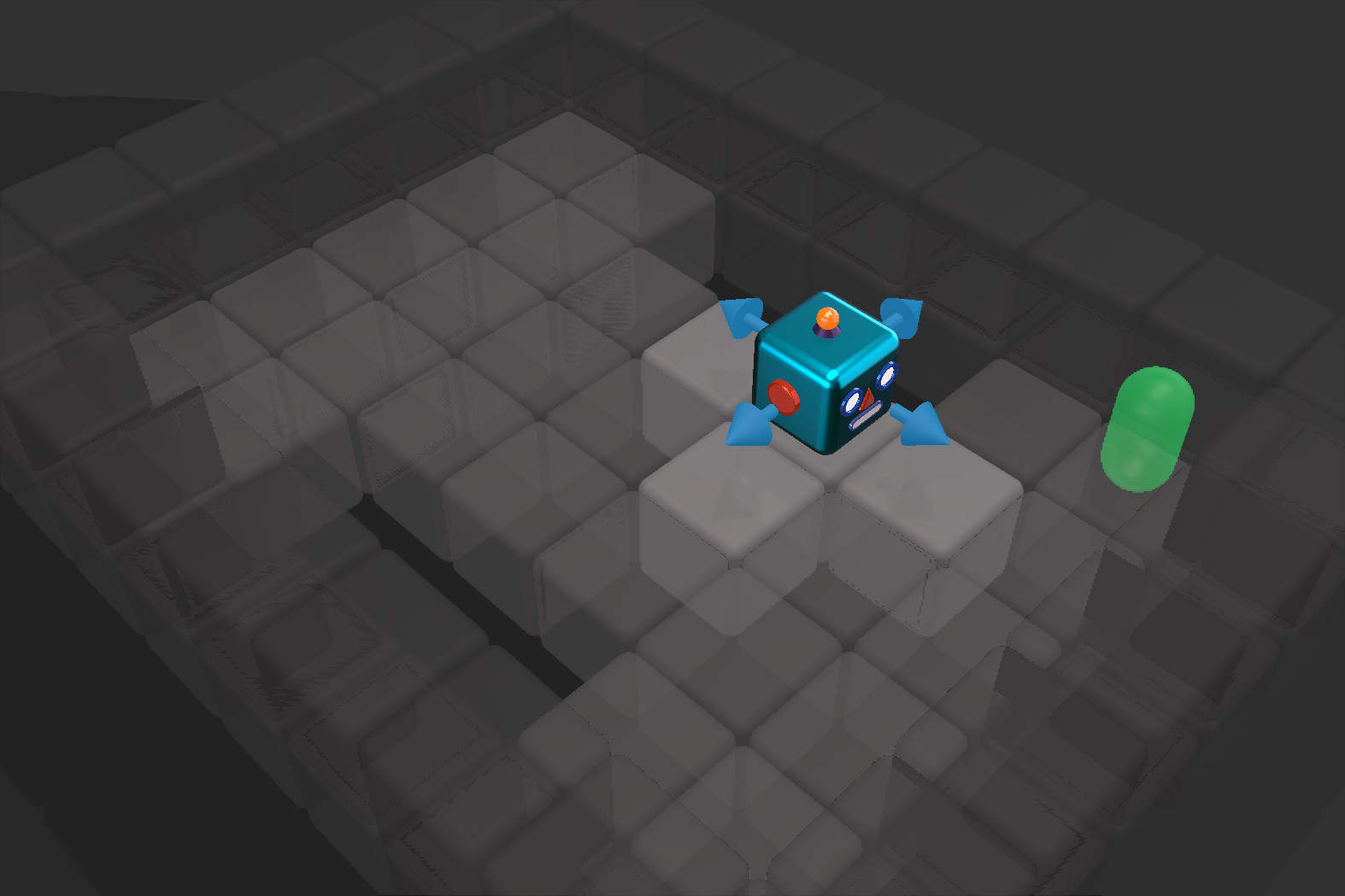}\label{fig:CROP:Action}
  }
  \subfloat[Object CROP]{
    \includegraphics[width=0.25\textwidth]{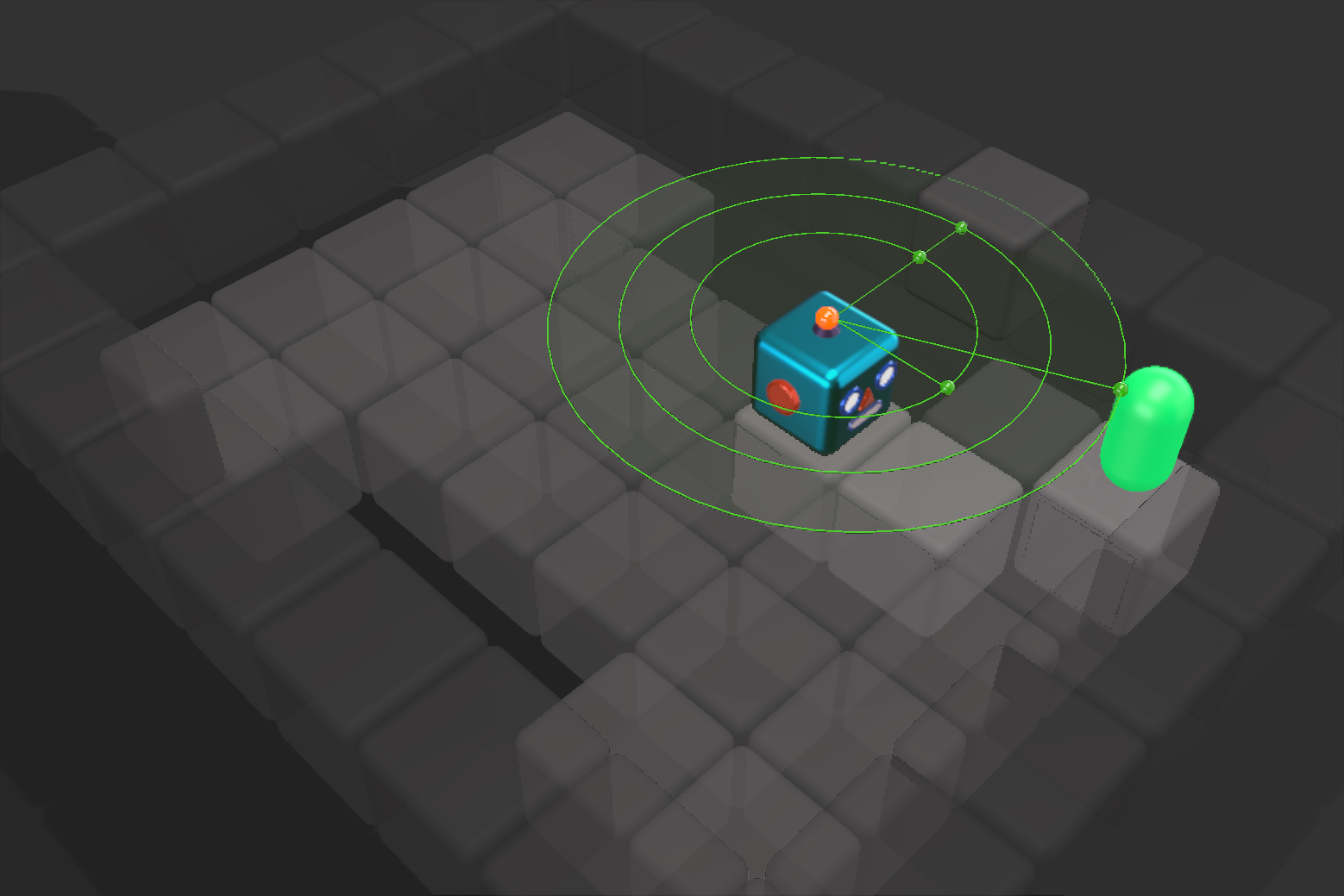}\label{fig:CROP:Object}
  }
  \subfloat[Full Observation]{
    \includegraphics[width=0.25\textwidth]{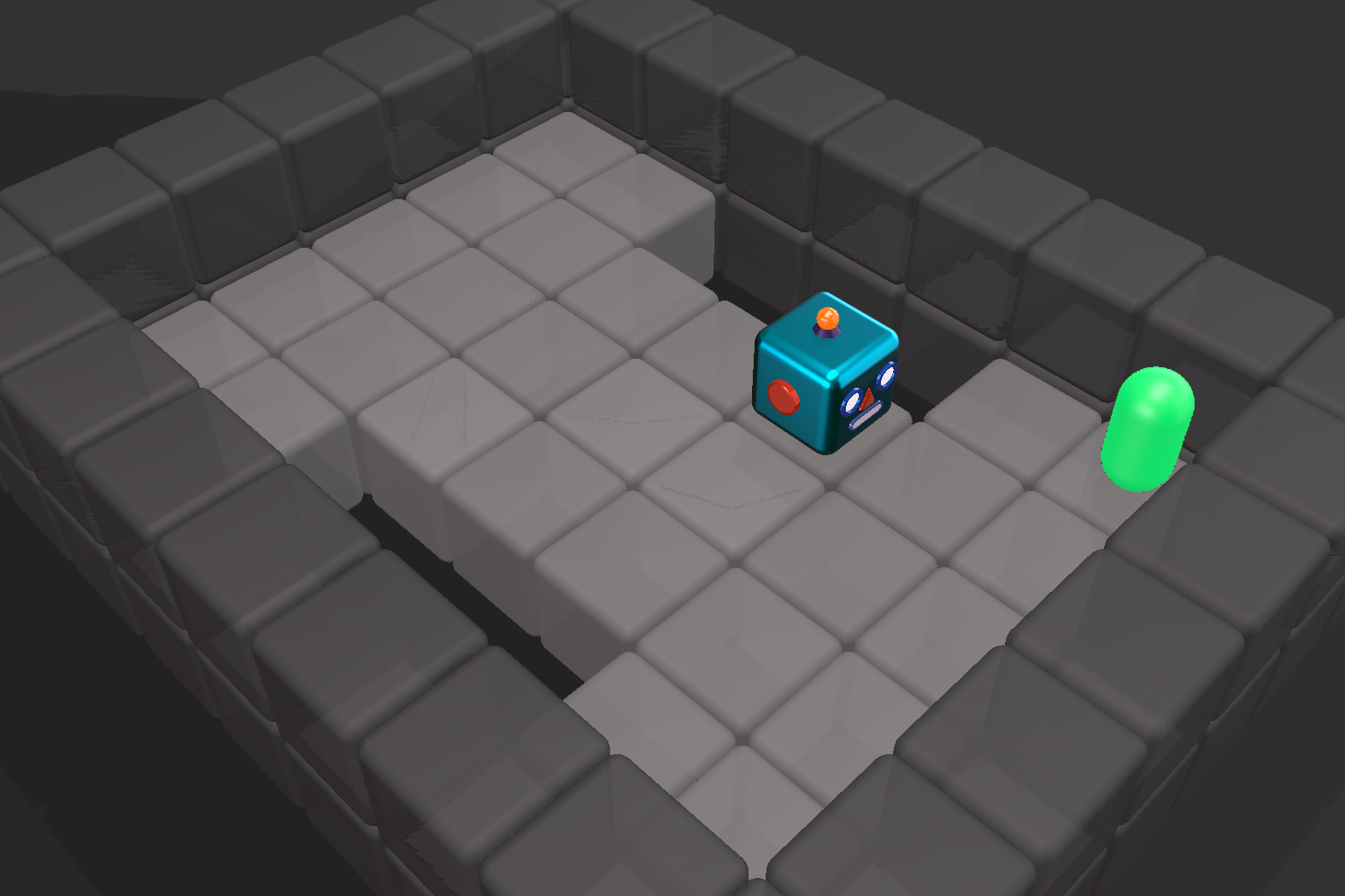}\label{fig:CROP:FO}
  }
    \caption{\textbf{CROP}: \textit{Compact Reshaped Observation Processing} based on the agent's position (blue / \autoref{fig:CROP:Radius}), action (light blue / \autoref{fig:CROP:Action}) and surrounding objects (green / \autoref{fig:CROP:Object}) in a fully observable (\autoref{fig:CROP:FO}) safety gridworld environment rewarding the attainment of the target (green).}
  \label{fig:CROP}
\end{figure*}

\newpage
\section{CROP}

To facilitate generalization to mechanics underlying the environment, we propose to reshape observations to a compact format containing information with specific relevance to the agent.
We argue, that if the reshaped state is invariant in similar situations, the policy optimization benefits from the more compact representation, while effects, previously described as benign overfitting \cite{bartlett2020benign}, can foster a policy that is robust regarding environmental changes. 

Formally, we suggest utilizing a reshaping function $\mathrm{CROP}: \mathcal{S} \mapsto \mathcal{S}^*$, where $\mathcal{S}^*$ is the reshaped observation space, similar to the observation function $\mathcal{O}$ in POMDPs.
In this paper, we use hard-coded compression functions, but we see great opportunities to extend this work to learning the compression functions as an approach to transfer learning in RL (meta-RL). By using the hard-coded compression function, we can leverage domain knowledge to accelerate learning. 
For interaction of the agent with the environment, CROP is used as surrogate observation, obtaining the modified observation $s^*_t = \mathrm{CROP}(s_t)$, where $s_t$ is the $d$-dimensional full observation of the environment at step $t$. 
\autoref{alg:CROP} demonstrates the application of CROP to an arbitrary policy optimization algorithm.

\begin{algorithm}[h]
    \caption{CROPed Policy Optimization}
    \textbf{Input}: Initialized Policy $\pi_\theta$\\
    \textbf{Parameters}: An observation processing function $\texttt{CROP}$,\\
    \-\quad\quad\quad\quad\quad \-\ \-\  A policy optimizer $\Theta$ and a learning rate $\lambda$\hfill
    \label{alg:CROP}
    \begin{algorithmic}[1] 
    \While{not done}                    \Comment{Collect Episode                       }
    \State $\tau \leftarrow \emptyset$  \Comment{Initialize empty rollout buffer }
    \For{step}                     \Comment{Perform Rollout                      }
        \State $s_t^* = \texttt{CROP}(s_t)$                             \Comment{CROP Observation}
        \State $a_t \sim \pi_\theta(a_t\mid s_t^*)$                     \Comment{Sample action         }
        \State $r_t = \mathcal{R}(s_t, a_t)$                            \Comment{Receive reward       }
        \State $s_{t+1} \sim \mathcal{T}(s_{t+1} \mid s_t,a_t)$              \Comment{Execute action        }
        \State $\tau \leftarrow \tau \cup \{s_t, a_t, r_t, s_{t+1}\}$   \Comment{Store transition       } 
    \EndFor
    \State $\theta \leftarrow \theta + \lambda \cdot \Theta(\tau)$         \Comment{Update Policy}
    
    \EndWhile  
    \end{algorithmic}
\end{algorithm}

To illustrate the proposed CROP mechanisms, we show their impact to an exemplary state in the fully observable safety training gridworld (cf. \autoref{fig:CROP:FO}), that is introduced in full detail in \autoref{sec:envs}.
However, it should be noted, that the proposed methods reflect basic concepts to asses the impact of CROP that may be refined, combined, or methodically applied to more complex observations. 
Concretely we propose the following three CROPs, visualized exemplary in \autoref{fig:CROP}: 
\begin{enumerate}
  \item \textbf{Radius CROP} (cf. \autoref{fig:CROP:Radius}): Reshapes the\\ observation to a $\rho$-sized radius around the agent:
    \begin{equation}
        CROP(s)^{Radius}_\rho = s_t[\alpha-\rho:\alpha+\rho] 
    \end{equation}
    \textit{State relevance:} ensured by positional proximity \\
    \textit{Required Information}: the $d$-dimensional position of the agent $\alpha$ and a padding character to produce consistent-sized observations on the edges
  \item \textbf{Action CROP} (cf. \autoref{fig:CROP:Action}): Reshapes the observation to states accessible via the immediate actions $\mathcal{A}$: 
    \begin{equation}
        CROP(s_t)^{Action}_{\alpha,\mu} = (s_t[\alpha+\mu_0], \dots, s_t[\alpha+\mu_n])
    \end{equation}
    \textit{State relevance:} ensured by state interactability and proximity, similar to Radius CROP, but sparser \\
    \textit{Additional Information}: the agent position $\alpha$ and a set of $n$ action-mutations $\mu$, assigning each possible action a $d$-dimensional mutation of the agent's position. 
  \item \textbf{Object CROP} (cf. \autoref{fig:CROP:Object}):  Reshapes the full observation (observation of every cell) to the distance vectors from the agent to the nearest $\eta$ cells for each object type $O \subset \mathcal{F}$.
    Thus, the resulting observation will have the dimension $dim(s^*_t)=(\lvert O\rvert \cdot \eta,d)$
    \begin{equation}\begin{aligned}
        CROP(s_t)_{O,\eta, \alpha, \sigma}^{Object} = \\(o_i - \alpha \quad \forall o \in O  \forall o_i \in \sigma(s_t, \alpha, \eta, o))
    \end{aligned}\end{equation}

    \textit{State relevance:} ensured by object interactability and proximity, comparable to a LIDAR sensor\\
    \textit{Additional Information:} A scan mechanism $\sigma(s,\alpha,\eta,o)$ to find the absolute position of $\eta$-nearest cells containing object type $o$. 
    
\end{enumerate}
All proposed methods transform the observation into a relative state centered/based around the agent, in contrast to the absolute, global, full observation $s_t$.
Radius CROP and Action CROP can be understood as explicit (hard encoded) attention mechanisms.
While all proposed methods transform the given MDP into an POMDP, Radius CROP is the closest resemblance of partial observability in a classical sense within a gridworld scenario, that are typically centered around the agent, especially in arbitrary-sized, potentially infinite environments. 

\section{Experimental Setup}
All implementations for the following evaluations can be found here \footnote{\url{https://github.com/philippaltmann/CROP}}.

\begin{figure}[t!]
  \centering
  \subfloat[Training $(-150,42)$]{
    \includegraphics[width=0.5\linewidth]{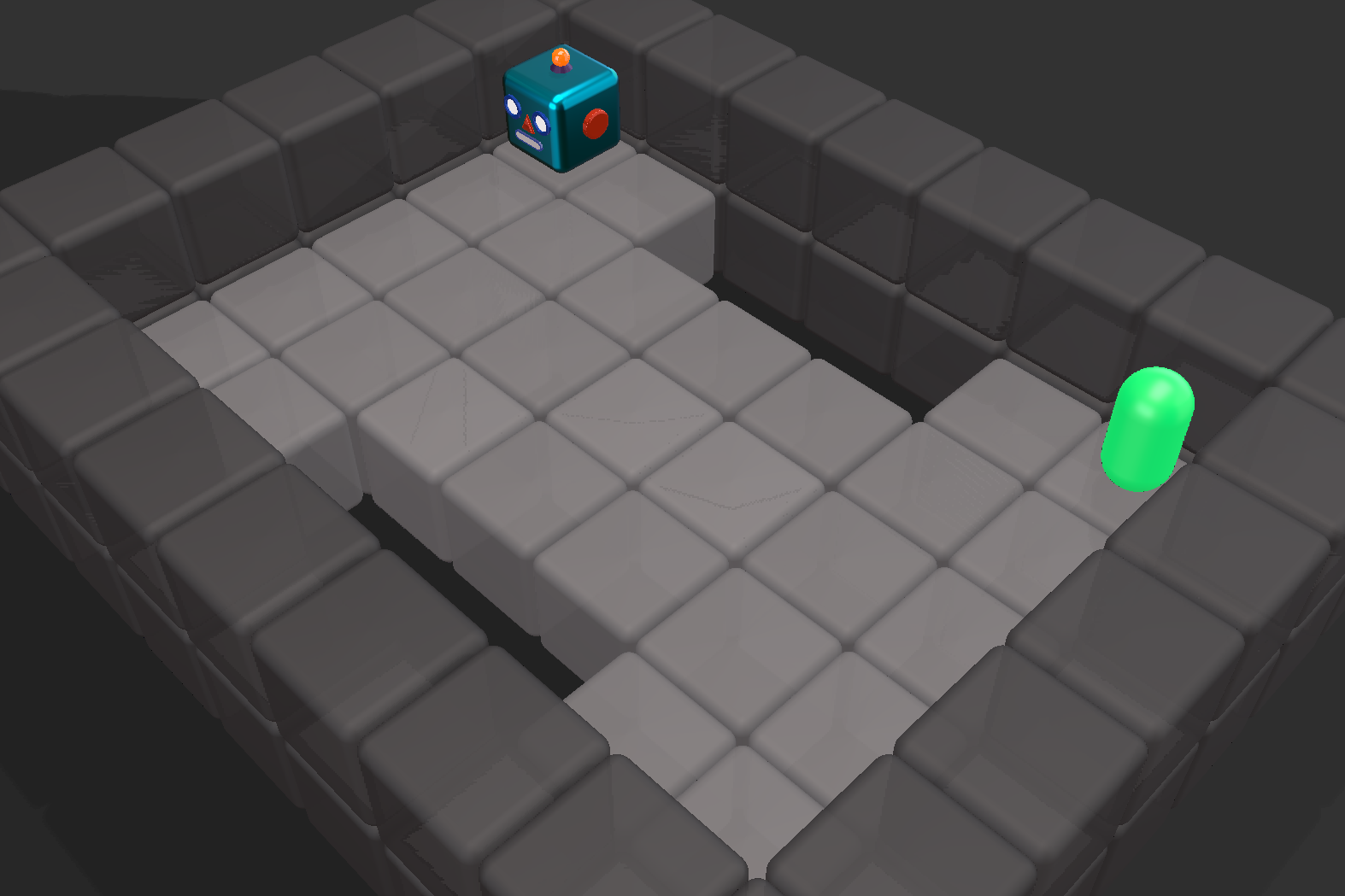}\label{fig:env:train}
  }
  \subfloat[Test $(-150,40)$]{
    \includegraphics[width=0.5\linewidth]{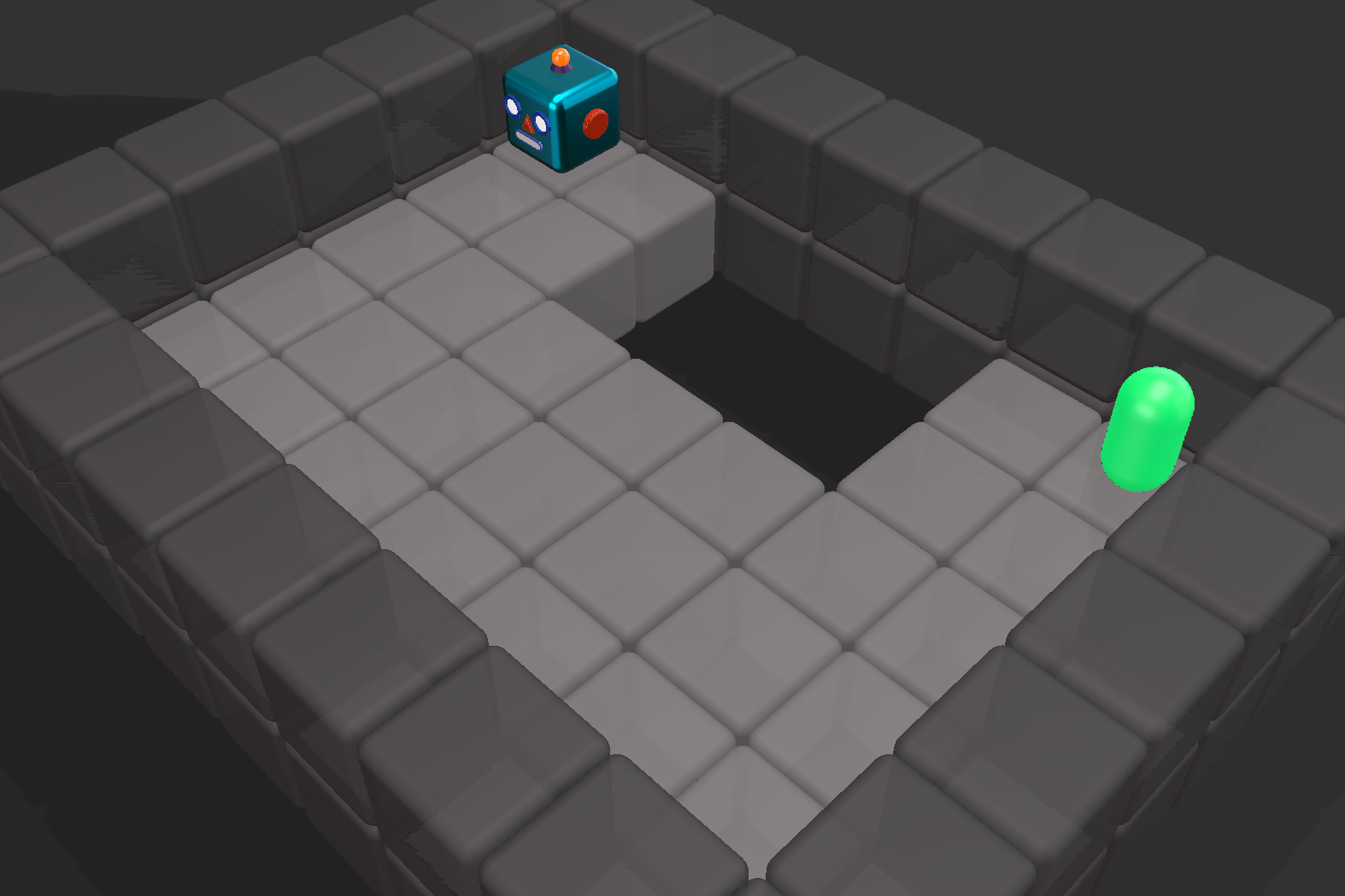}\label{fig:env:test}
  }\\
  \subfloat[Single Maze-7 $(-100,42)$]{
    \includegraphics[width=0.5\linewidth]{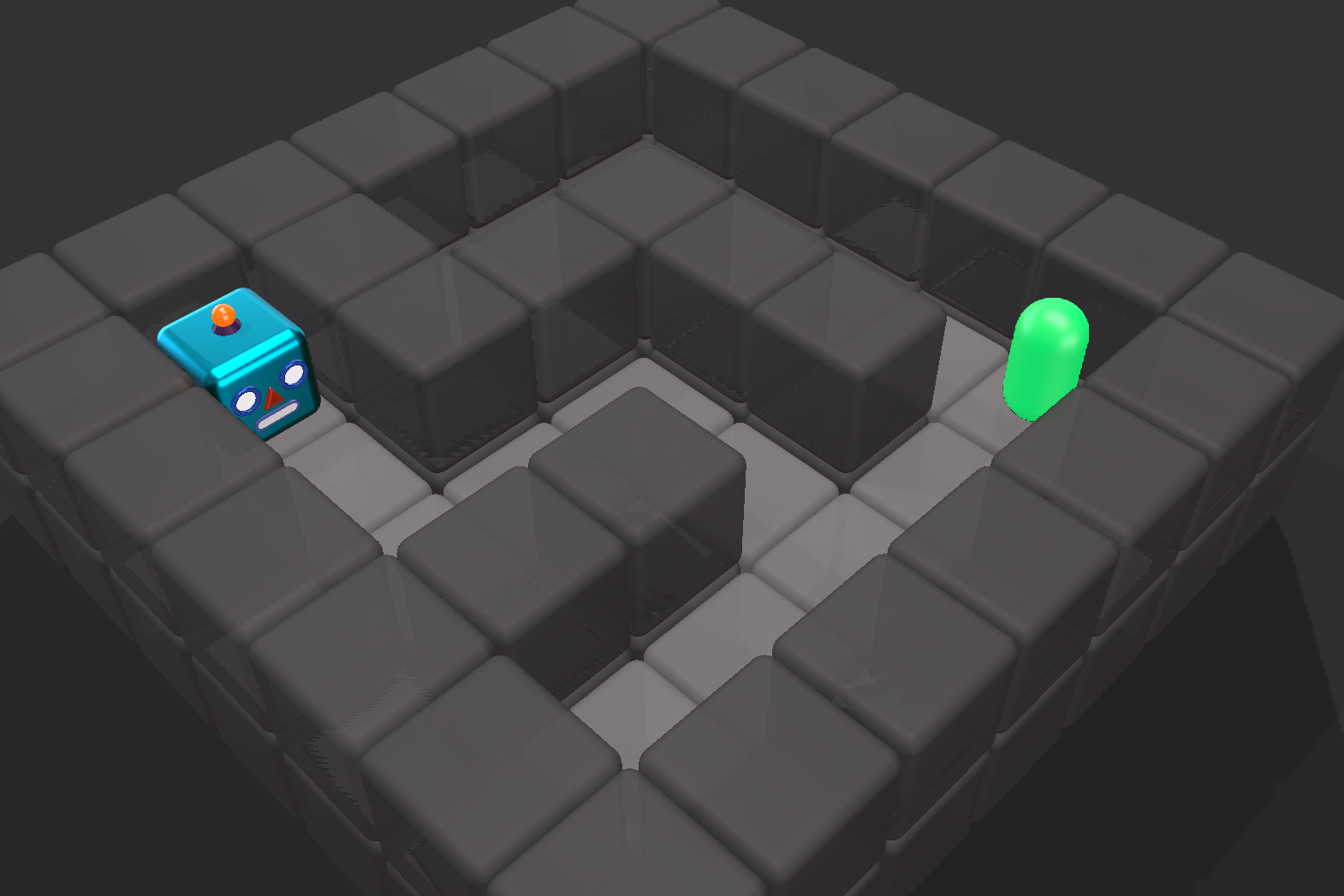}\label{fig:env:maze7}
  }
  \subfloat[Single Maze-11 $(-100,30)$]{
    \includegraphics[width=0.5\linewidth]{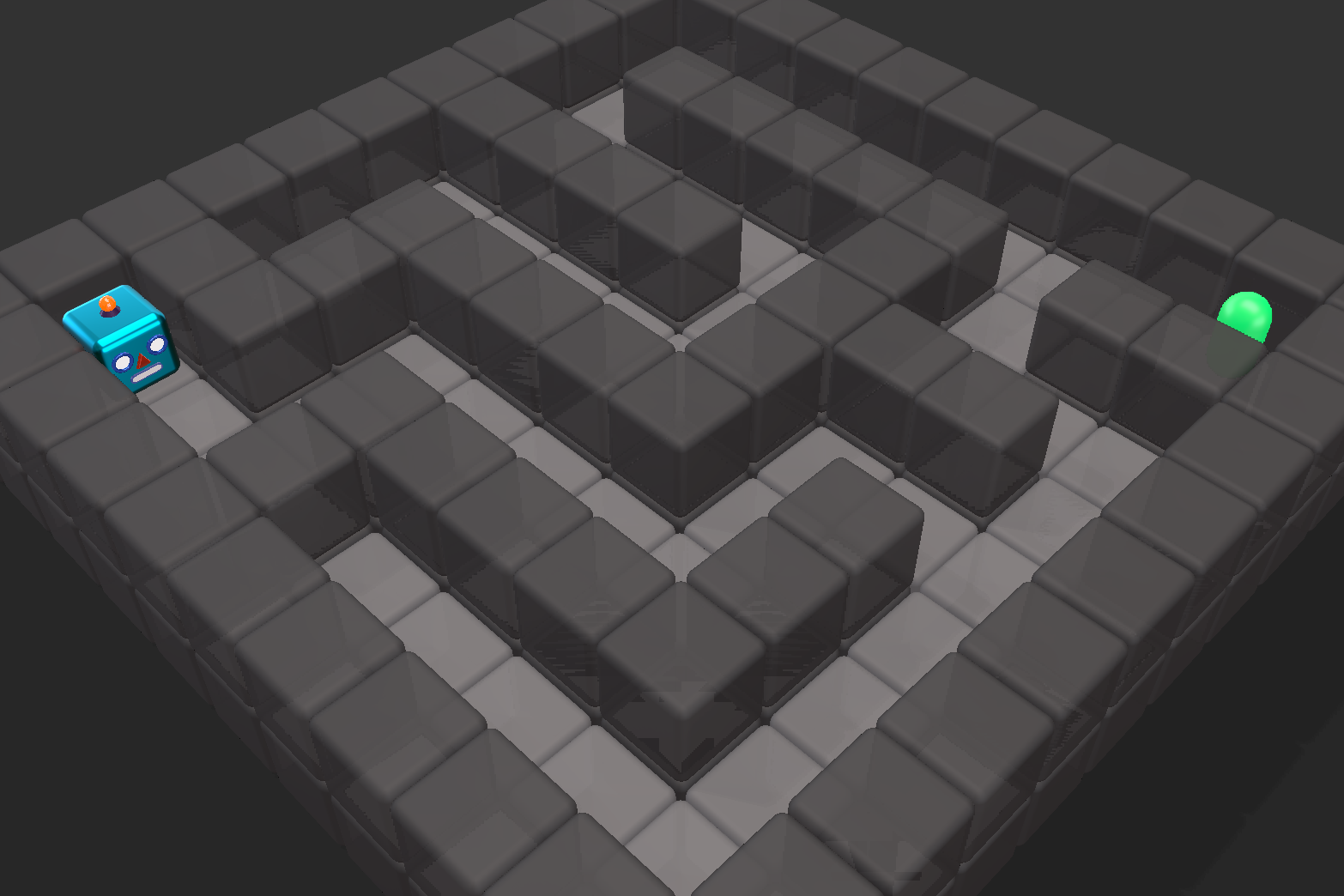}\label{fig:env:maze11}
  }
\caption[Evaluation Environments]{Evaluation environments and reward ranges: Holey distributional shift gridworlds inspired by \cite{leike2017ai} for training (\autoref{fig:env:train}) and shifted evaluation (\autoref{fig:env:test}), as well as deterministic configurations of Maze 7 (\autoref{fig:env:maze7}) and Maze11 (\autoref{fig:env:maze11}) inspired by \cite{cobbe2020leveraging}. The agents' goal is to reach the target (green capsule), rewarded with $50$. To incentivize the shortest path, every step is penalized with $-1$. Holes immediately terminate the episode and are penalized with $-50$. Episodes are terminated after a maximum of $100$ steps.}\label{fig:envs}
\end{figure}

\paragraph{Environments:}\label{sec:envs}
To provide poof-of-concept for CROP we used two holey safety gridworlds inspired by \cite{leike2017ai} comprising an $(7,9)$ observation space with a set of five discrete features $\mathcal{F}=\{Wall, Field, Hole, Goal, Agent\}$ and and four discrete actions $\mathcal{A}=\{Up, Right, Down, Left\}$.
To specifically asses the models' robustness to changes in the environment, we train all models in the single training configuration visualized in \autoref{fig:env:train} and test their performance in the unseen distributionally shifted environment shown in \autoref{fig:env:test}.
For further evaluation and comparisons in \autoref{sec:bench} we use $(7,7)$- and $(11,11)$-sized generated mazes inspired by \cite{cobbe2020leveraging} with an identical action space and the reduced set of fields $\mathcal{F}=\{Wall, Field, Goal, Agent\}$ shown in \autoref{fig:env:maze7} and \autoref{fig:env:maze11} respectively.
Again, to assess generalization, unseen configurations were used to test the trained policies. 
Therefore, we use a pool of 100 randomly generated mazes explicitly excluding the deterministic configuration to test polices trained in the single maze configurations and to train policies that are tested in the single deterministic environment.
The reward range of randomly generated mazes is dependent on the shortest path, thus variable. 
However, the hardest possible mazes yielding the longest possible shortest paths, result in an optimal reward of 34 an 2 for Maze-7 and -11 respectively. 
To increase training speed, we trained all policies using four parallel environments. 

\begin{figure*}[!t]
  \centering
  \subfloat[Validation Return in Training Environment]{
    \includegraphics[width=0.4\textwidth]{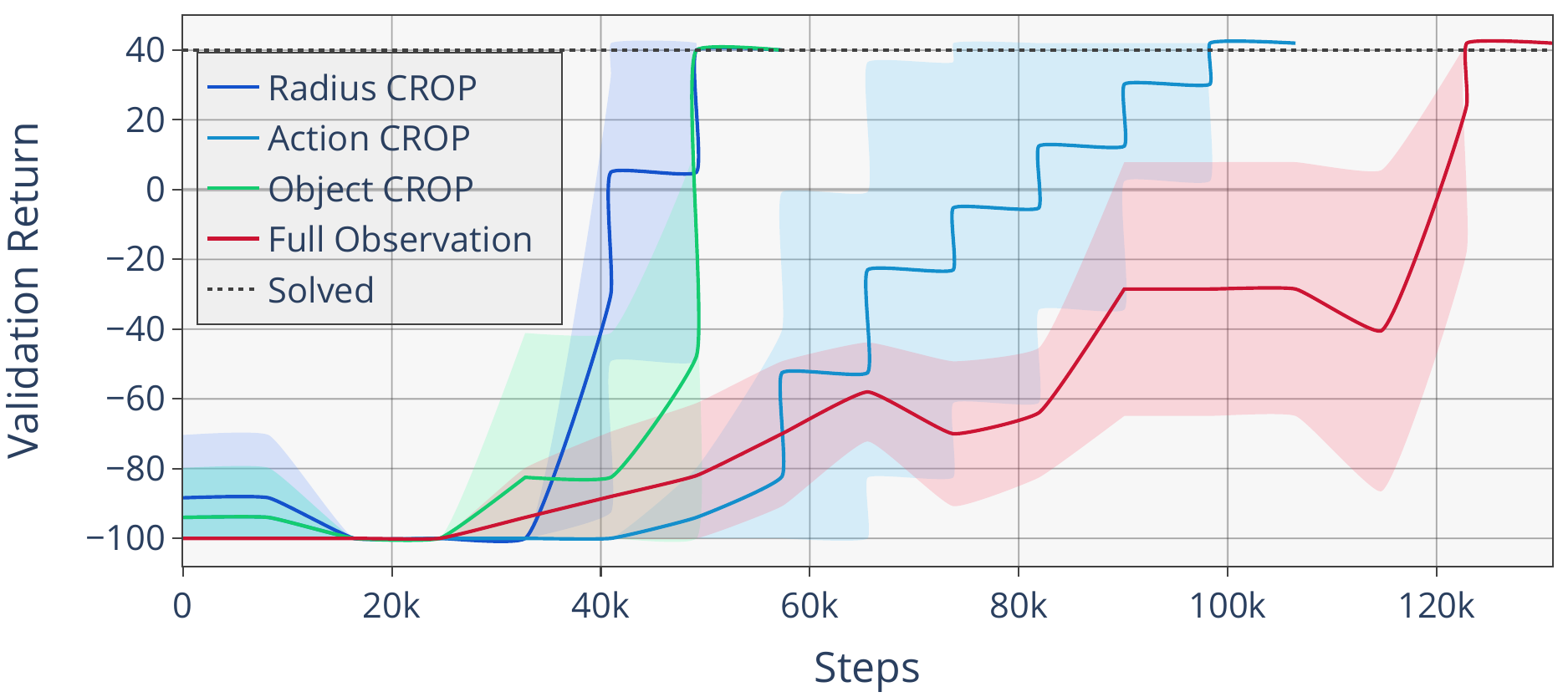}\label{fig:eval:train}
  }
  \subfloat[Evaluation Return in Test Environment]{
    \includegraphics[width=0.4\textwidth]{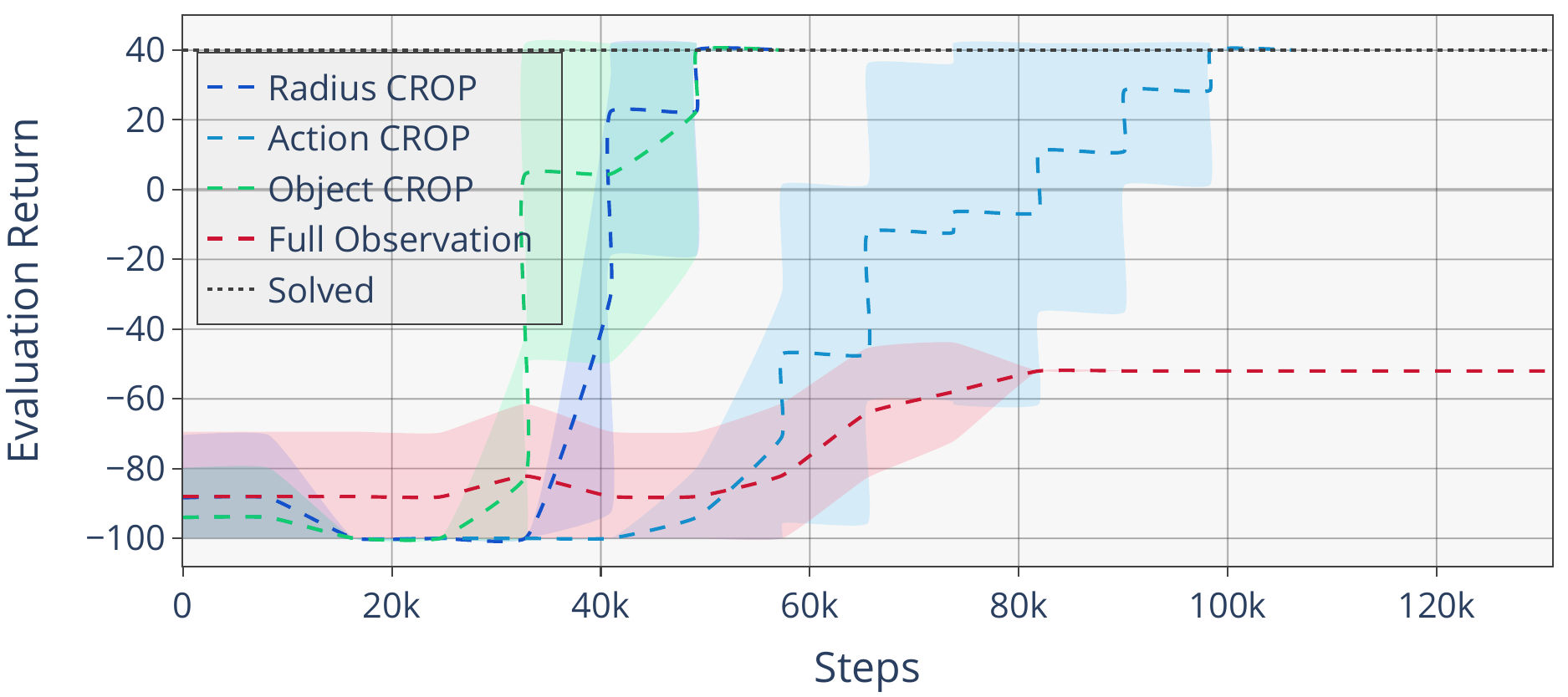}\label{fig:eval:test}
  }
  \caption{\textbf{CROP Evaluation: } Comparing Radius- (blue), Action- (light blue), Object-CROP (green), and Full Observability (red) in the Distributional Shift Safety Environment. The number of steps taken in the environment is on the x-axis and the Validation Return (solid lines in \autoref{fig:eval:train}), and Evaluation Return (dashed lines in \autoref{fig:eval:train}) on the y-axis averaged over eight random seeds. The shaded areas mark the 95\% confidence intervals. The reward threshold of 40 is displayed by the dotted line. 
  } \label{fig:eval}
\end{figure*}

\paragraph{Policy Optimization: }
Whilst applicable to any policy optimization algorithm, we chose to evaluate CROP as an extension to PPO, having shown to be a robust and universally applicable state-of-the-art choice \cite{schulman2017proximal}.
We furthermore built upon the implementations by \cite{stable-baselines3}, extending upon \cite{gym}.
To analyze the effect of CROP and demonstrate its strengths we provide ablation studies and compare training and evaluation using \textit{Full Observations} (FO) to all three CROPed observations in \autoref{sec:eval}. 
Furthermore we provide benchmark comparisons to a FO \textit{Advantage Actor Critic} (A2C) from \cite{stable-baselines3} and an alternative method for improved generalization, data augmentation, in \autoref{sec:bench}.

\paragraph{Data Augmentation (RAD): }
To provide a state-of-the-art comparison, we implemented data augmentation mechanisms for reinforcement learning (RAD) according to \cite{laskin2020reinforcement} and evaluate their impact in randomly generated mazes in \autoref{sec:bench}.
For a fair comparison to CROP however, we also apply the proposed methods to the discrete full representation of the environment, instead of using images, and use PPO for training as suggested by the authors.
Therefore, all image-based transformations like grayscale ore color-jitter are not applicable. 
Thus, we use random-crop, -translate, and -cutout with the same ratios suggested by \cite{laskin2020reinforcement}, causing the full observation to be randomly cropped to (6,6) and (9,9), randomly translated to the original shape (7,7) and (11,11), and cut-out by patches sized in the ranges (0,2) and (0,3) in all dimensions.

\paragraph{Hyperparameters:}
For training PPO, we adopted the default parameters implemented by \cite{stable-baselines3}, also suggested in \cite{schulman2017proximal}.
For the safety environments, we trained all models for a maximum of 1M steps, terminating once 99\% of the optimal return (40 for the training environment) is reached.  
For \textbf{Radius CROP} we set the radius $\rho = (2,2)$, resulting in an observation shape of $dim(s^*_t)=\rho\cdot2+1 = (5,5)$, padded with wall fields. 
Given the four possible actions $\mathcal{A}=\{Up,Right,Down,Left\}$ we parameterized \textbf{Action CROP} with $\mu=[(-1,0),(0,1),(1,0),(0,-1)]$, resulting in an observation shape of $dim(s^*_t)=\lvert\mathcal{A}\rvert=(4)$. 
Regarding \textbf{Object CROP} we chose $\eta=1$ for all safety environments and $\eta=2$ for all mazes and the set of objects to be detected to be all possible objects excluding the agent itself: $O=\mathcal{F}\setminus\{Agent\}$, resulting in $O=\{Wall,Field,Hole,Goal\}$ and the observation shape $dim(s^*_t)= (4,2)$ for the train and test environments (cf. \autoref{fig:env:train} and \autoref{fig:env:test}), as well as $O=\{Wall,Field,Goal\}$ and the observation shape $dim(s^*_t)=(3,2)$ for all maze environments (cf. \autoref{fig:env:maze7} and \autoref{fig:env:maze11}).

\paragraph{Metrics:}
To reflect both the training performance and the generalization capabilities, we regularly (every $2^{13}$ steps) evaluated the policies during training (without further training) in both the training and an unseen test configuration, reflected by the metrics \textit{Validation Return} and \textit{Evaluation Return} respectively. 
The return is either averaged over 100 non-deterministic episodes (for all maze configurations) or based on a single deterministic episode (for all holey environments, to reflect the certain safety of the current policy).
Additionally, to increase significance, all runs are averaged over eight independent seeds.

\section{Evaluation}\label{sec:eval}

To provide proof of concept, that the reduced information is sufficient for learning an optimal policy, the following section contains ablation studies comparing the performance of PPO for learning a policy in the holey safety training environment (\autoref{fig:env:train}) using Full Observations (FO), Object CROP (O-CROP), Action CROP (A-CROP) and Radius CROP (R-CROP) (cf. \autoref{fig:CROP}).

The progress of the Validation Return throughout training in the holey safety environment (\autoref{fig:env:train}) is shown in \autoref{fig:eval:train}.
Overall, all compared approaches find optimal policies, reaching returns above the solution threshold 40, within 150k steps. 
However, presumably caused by an insufficient representation, finding an optimal policy is the slowest using FOs. 
On the other hand, the results for all CROPs show, that the compressed information serves sufficient for training an optimal policy. 
Furthermore, caused by the increased relevance of observed states, the required training steps within the environment are reduced by 50\% for R-CROP and O-CROP, compared to FO.  
The comparably slower training observed for A-CROP is probably caused by its too sparse observation, containing only the four neighboring states, which, again hinders fast convergence. 

However, the real benefits of CROP are exposed in the Evaluation Return shown in \autoref{fig:eval:test}.
Caused by the previously unseen shift of the positions of the holes in the environments, polices trained using FO only reach returns up to $-50$. 
This return is most likely caused by polices that are terminated by a hole, instead of the agent reaching the target state. 
On the other hand, all CROP-trained polices show to be robust to said shift, resulting in significantly increased Evaluation Returns. 
Comparing the Evaluation Return with the Validation Returns in \autoref{fig:eval:train} shows that said robustness is obtained in parallel to learning to solve the training environment, manifesting the advantages of CROP.
Furthermore, all CROP trained polices reach Evaluation Returns above the solution threshold of the test environment of 38, resulting not only in a behavior that is able to navigate to the target state, but also uses the shortest possible path.

\begin{figure}[t]
  \centering
  \subfloat[FO Policy]{  
    \includegraphics[width=0.22\textwidth]{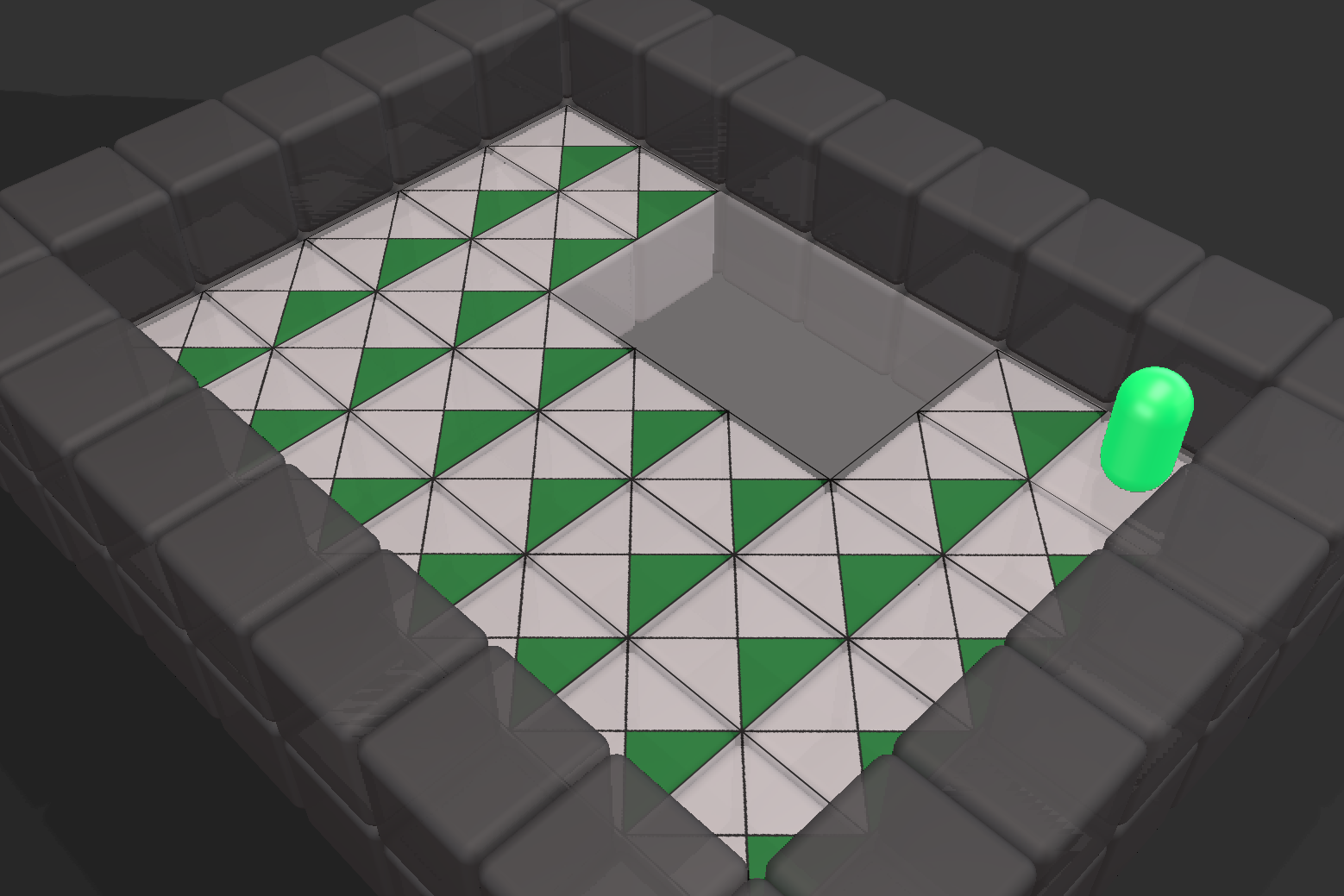}\label{fig:eval:hm:FO}
  }
  \subfloat[R-CROP Policy]{
    \includegraphics[width=0.22\textwidth]{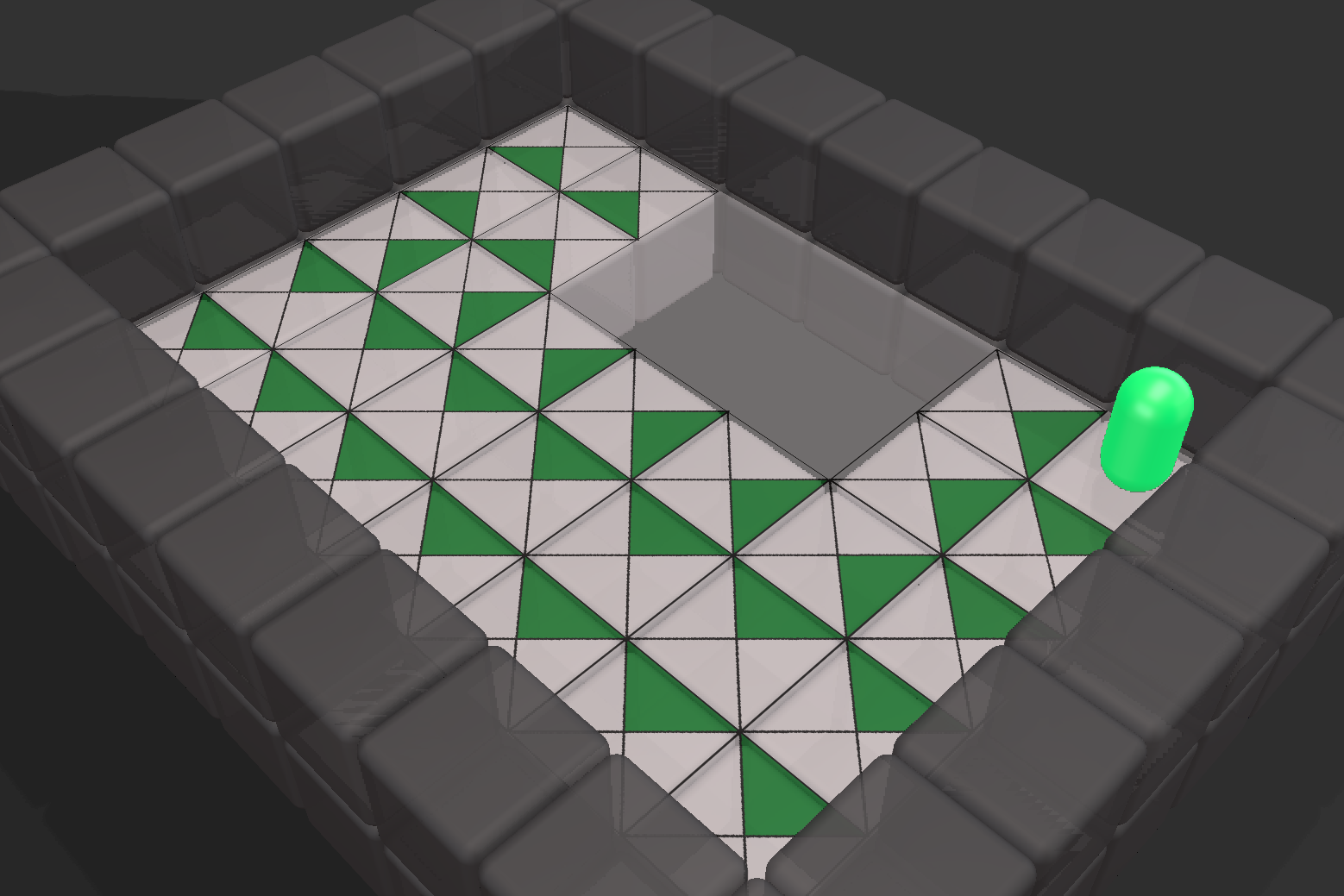}\label{fig:eval:hm:R-CROP}
  }
  \caption{\textbf{Validation Heatmaps} visualizing the dominant action of the PPO-trained policies using Full Observations (\autoref{fig:eval:hm:FO}) and Radius CROP (\autoref{fig:eval:hm:R-CROP}) for each possible state in the unseen shifted test environment (cf \autoref{fig:env:test})}\label{fig:eval:heatmaps}
\end{figure}

\autoref{fig:eval:hm:FO} and \autoref{fig:eval:hm:R-CROP} provide further insights into the resulting policies, showing a heatmap visualization of the dominant (deterministic) action in each state of the unseen test environment, chosen by the FO- and R-CROP-trained policies respectively. 
Heatmaps depicting A-CROP and O-CROP policies are omitted, as their behavior resemble the results shown for R-CROP.
As assumed above FO-trained policies reveal a behavior directly navigating into the nearest hole, even though the policy has learned to evade the holes in the training environment (cf. \autoref{fig:eval:hm:FO}). 
This behavior suggest that the trained policies overfit to the full observation of the training environment. 
The polices trained using R-CROP on the other hand are able to evade the shifted holes, even though, the environment has not been seen during training (cf. \autoref{fig:eval:hm:R-CROP}). 
Furthermore, the heatmap reveals, that the trained policy is able to reach the target within the shortest possible path, from any position, even though, the training was only conducted with the agent starting in the top left field. 
This generalization capability can not be observed for the FO-trained policies at all, only reaching the target from the neighboring field and otherwise failing to fulfill the intended task. 

Overall, the evaluations results provide evidence that CROP reduces the information to an efficient representation containing the important information for finding an optimal policy, whilst accelerating training performance and improving robustness to distributional shifts by the removal of unimportant details, otherwise prone to overfitting.

\begin{figure*}
\begin{minipage}{0.33\textwidth}\label{fig:bench:mazes7}
  \centering
  \subfloat[Validation Return in Random Maze-7]{
    \includegraphics[width=\textwidth]{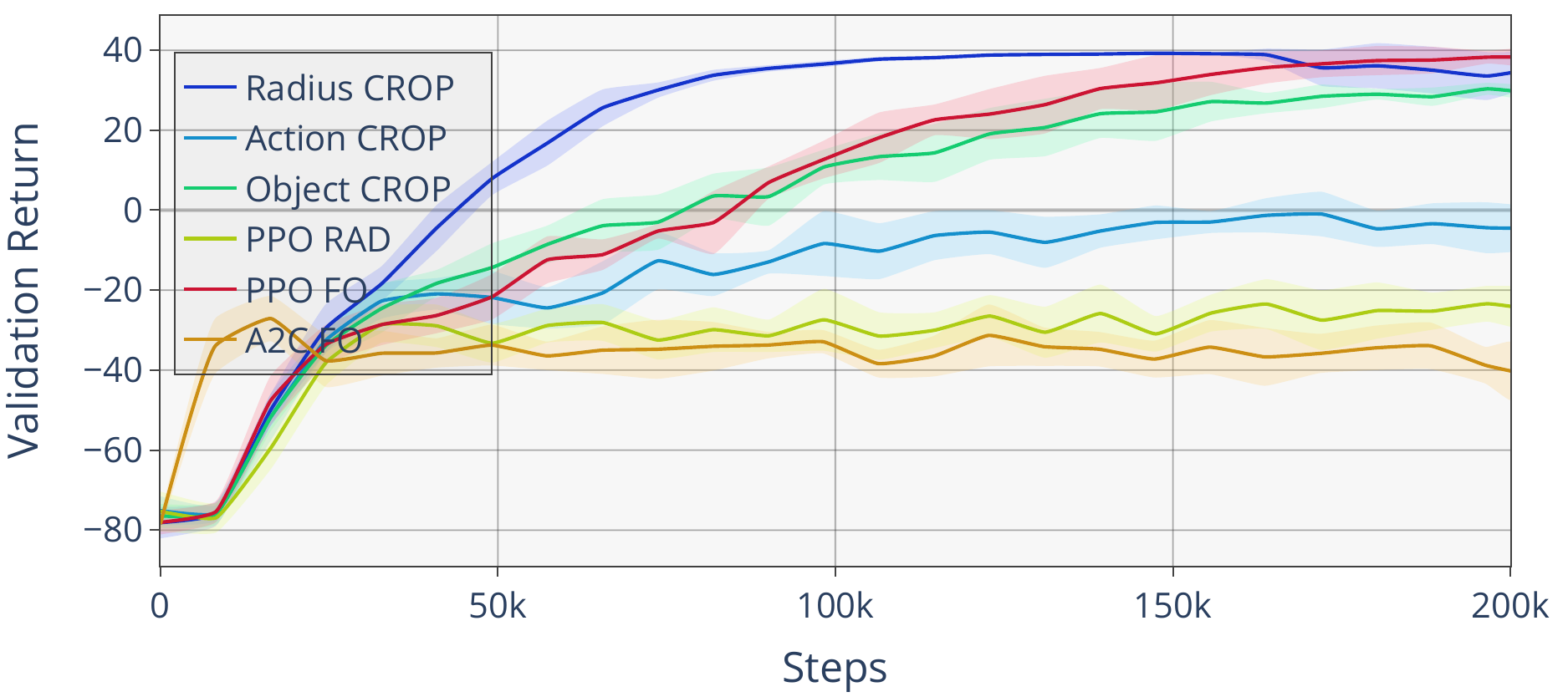}\label{fig:mazes7:train}
  }\\
  \subfloat[Evaluation Return in Single Maze-7]{
    \includegraphics[width=\textwidth]{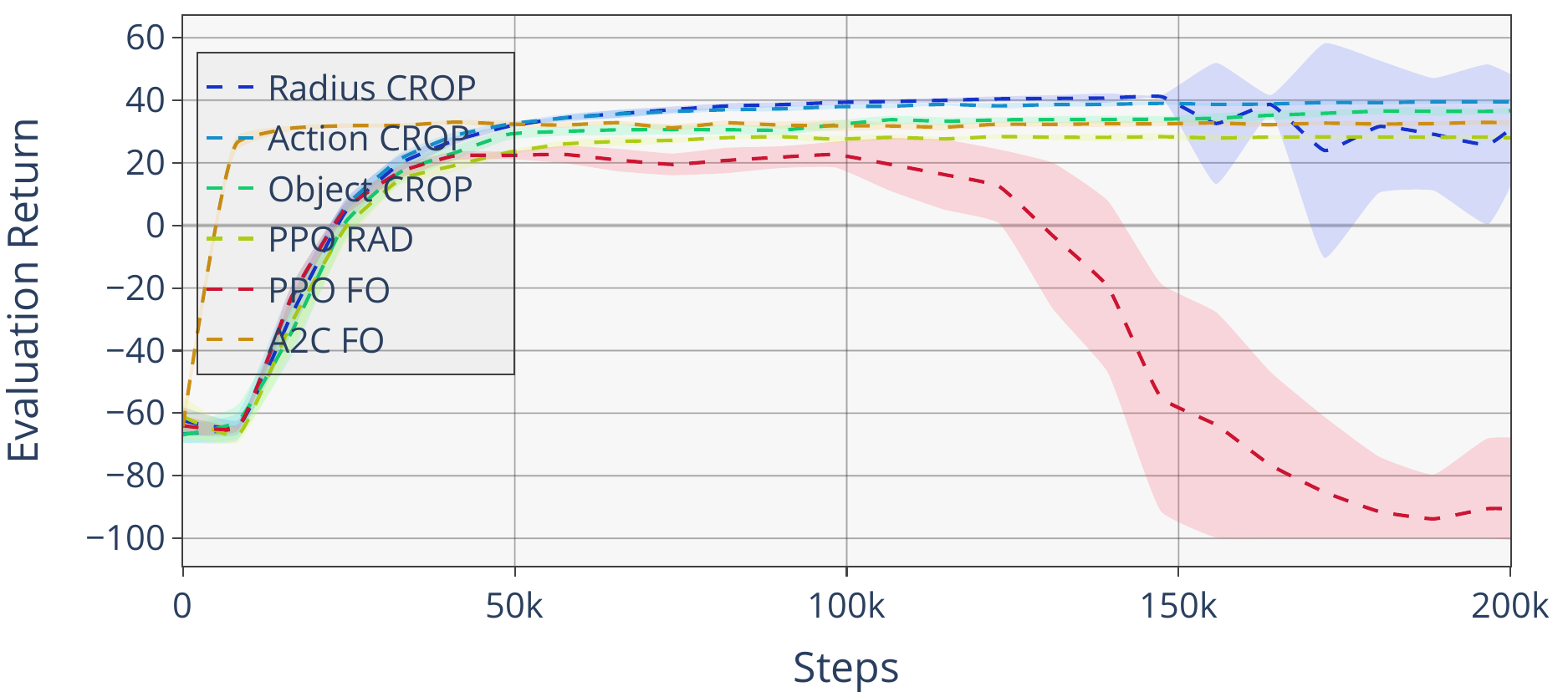}\label{fig:mazes7:test}
  }
\end{minipage}
\begin{minipage}{0.33\textwidth}
\centering
  \subfloat[Validation Return in Single Maze-11]{
    \includegraphics[width=\textwidth]{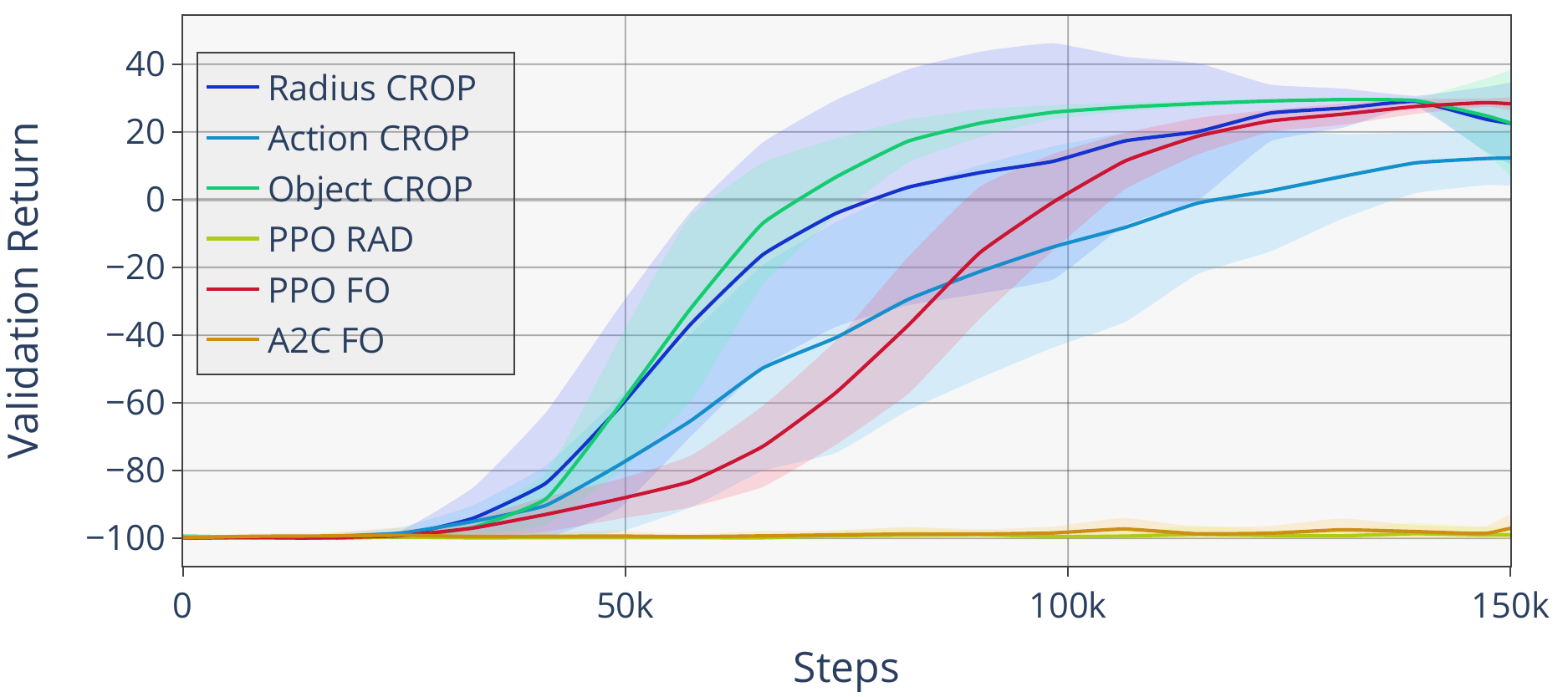}\label{fig:maze11:train}
  }\\
  \subfloat[Evaluation Return in Random Maze-11]{
    \includegraphics[width=\textwidth]{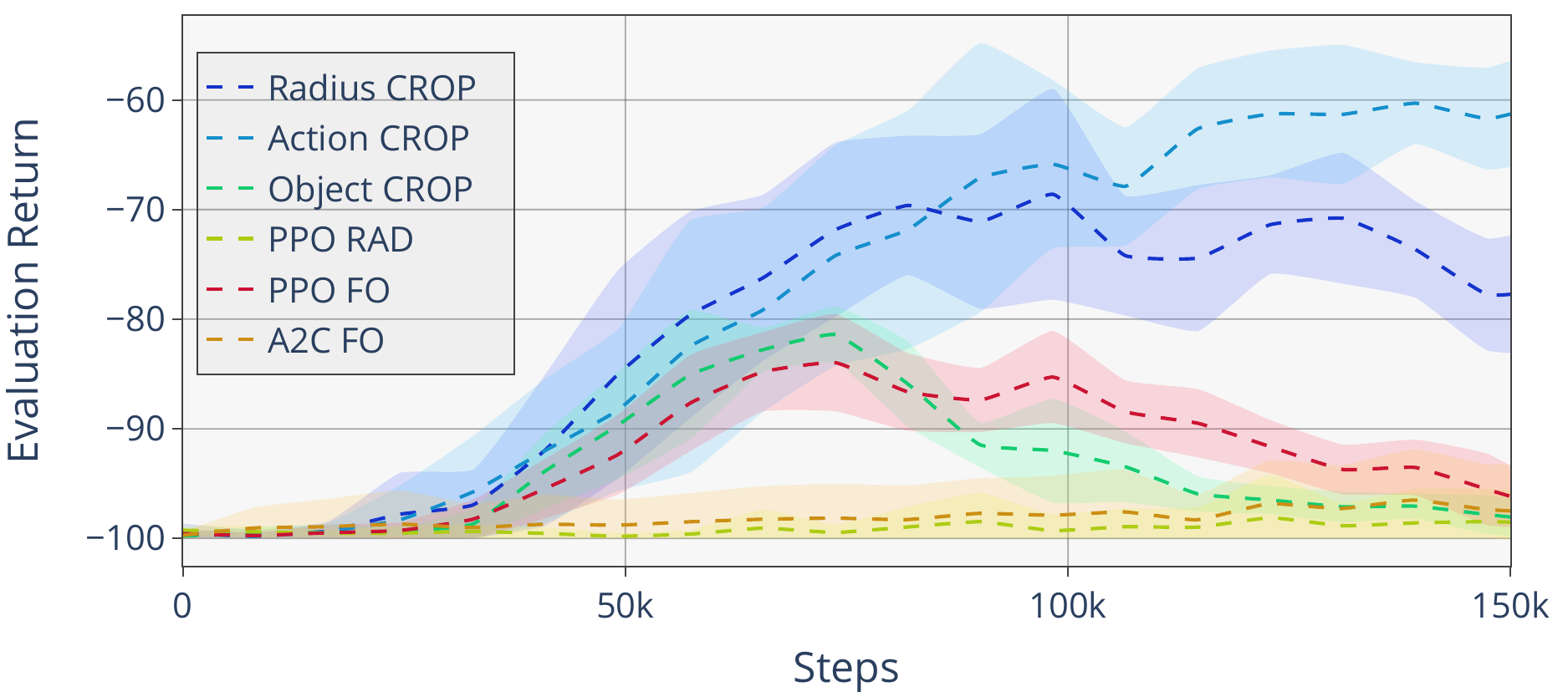}\label{fig:maze11:test}
  }
\end{minipage}
\begin{minipage}{0.33\textwidth}
\centering
  \subfloat[Validation Return in Random Maze-11]{
    \includegraphics[width=\textwidth]{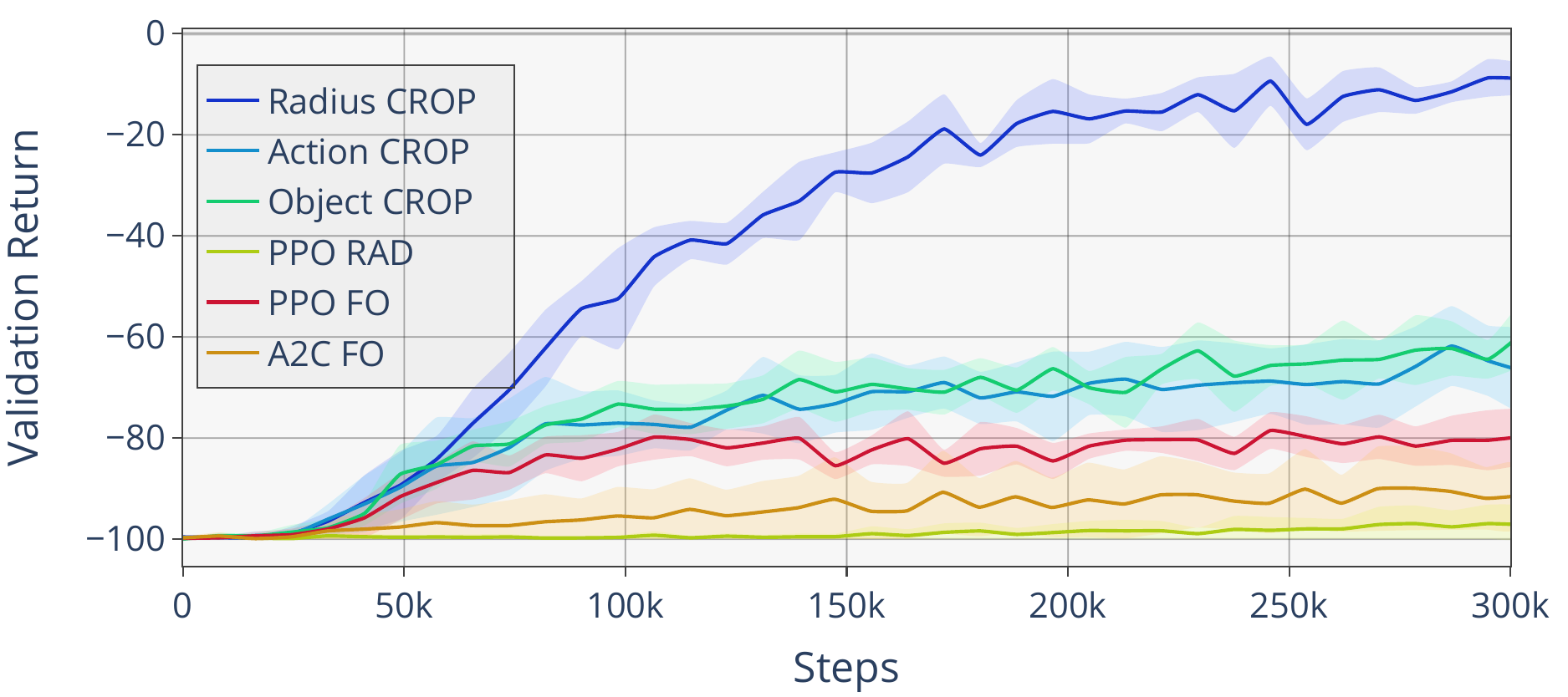}\label{fig:mazes11:train}
  }\\
  \subfloat[Evaluation Return in Single Maze-11]{
    \includegraphics[width=\textwidth]{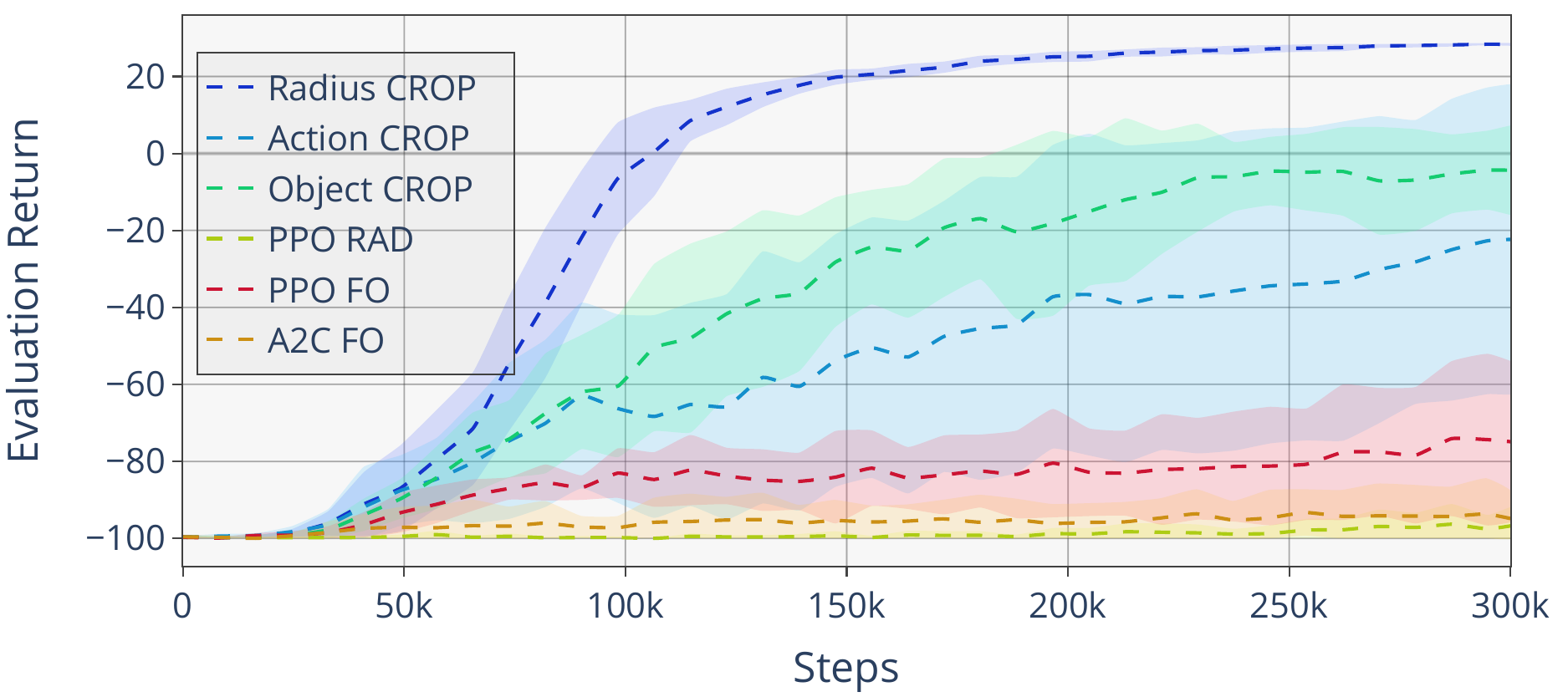}\label{fig:mazes11:test}
  }
\end{minipage}
\caption{\textbf{Generalization Benchmark:} Comparing R-CROP (blue), A-CROP (light blue), O-CROP (green), RAD (yellow) and FO (red) for training PPO, and A2C using FO (orange) in 100 random Maze-7, a single Maze-11, and 100 random Maze-11 configurations for 200k, 150k, and 300k steps (x-axis) respectively, w.r.t. the Validation Return (solid lines in \autoref{fig:mazes7:train}, \autoref{fig:maze11:train}, and \autoref{fig:mazes11:train}) and the Evaluation Return (dashed lines in \autoref{fig:mazes7:test}, \autoref{fig:maze11:test}, and \autoref{fig:mazes11:test}) on the y-axis, averaged over eight random seeds. The shaded areas mark the 95\% confidence intervals.} 
  \label{fig:bench}
\end{figure*}

\section{Benchmark Comparison}\label{sec:bench}

To further asses the generalization capabilities of CROP-trained policies, we provide comparisons to the Full Observation (FO) for training both PPO and A2C and Augmented Observations (RAD) for training PPO in three increasingly complex maze settings. 
As suggested in \cite{cobbe2020leveraging}, we first evaluate the generalization performance when training for 200k steps using a pool of randomly generated configurations. 

The Validation Return is shown in \autoref{fig:mazes7:train}. 
Similar to the previous evaluations, all observation types provide sufficient information for learning an expedient policy using PPO. 
However, both FO-A2C and RAD only reach performances below -20, while both FO-PPO and R-CROP reach near optimal performance of around $40$.
Again, presumably caused by the compressed observation, R-CROP shows the fastest convergence within about 100k steps, where FO and O-CROP increase the required steps by about 50\%, 
A-CROP also provides sufficient information to reach the target, but its sparsity seems to hinder convergence to an optimal behavior. 

However, evaluating the polices' generalization capabilities by analyzing their performance in the single unseen Maze-7 configuration, shown in \autoref{fig:mazes7:test}, all CROP-trained polices show near optimal performance, with final Evaluation Returns of A-CROP, R-CROP, and O-CROP around 40. 
On the other hand, FO-trained policies initially show good generalization to the unseen test configuration within the first 100k steps.
However, as the trained policy explores a near optimal solution in the training configuration, reaching Validation Returns upwards of 20, generalization drastically decreases and nearly drops to the minimum of -100 after 200k steps, indicating evaluation episodes where the final policy is not able to reach the target at all. 
This artifact of overfitting not showing for all CROP methods again confirms above assumptions regarding benign overfitting, thus improved generalization, encouraged by the improved observation. 
Interestingly, both FO-A2C and RAD-PPO show Evaluation Returns similar to CROP, for the single maze configuration, indicating either improved generalization, or, just less overfitting, given their Validation Return performed worst in comparison. 

Increasing the maze size to 11 whilst relaxing the training complexity by using only a single configuration, similar effects show (cf. \autoref{fig:maze11:train}).
All CROP-trained, as well as the FO-PPO trained policies reach near optimal performance within about 140k steps, while R-CROP and O-CROP show the fastest convergence. 
FO-A2C- and RAD-PPO-trained polices show the worst performance around the minimal return of -100. 
Interestingly however, even though showing slightly suboptimal training performance of around 15, A-CROP reaches the highest Evaluation Returns, thus best generalization. 
Note, that in this scenario, generalizing from a single training environment to 100 random unseen test environments is a notably harder challenge compared to the previous experiments, explaining the considerably lower overall Evaluation Returns. 
Nevertheless, while FO- and O-CROP-trained policies show indications of overfitting to the training environment in their Evaluation Return, R-CROP and A-CROP trained policies generalize to a behavior at least reaching the goal within unseen maze configurations. 
FO-A2C- and RAD-PPO-trained polices merely reach the target of the test environments at all, which is unsurprising given their training performance. 

Finally, \autoref{fig:mazes11:train} shows the Validation Return training in 100 random Maze-11 configurations. 
In contrast to the previous results, only Radius CROP-trained policies reach performances above -50, with the final return -10.
Remarking the theoretical worst case maximum return for generated Maze-11 configurations of 2, however, this translates to an 90\% optimal behavior. 
Furthermore, as for all previous experiments, the agent is not able to observe the target from the initial position using R-CROP, making this high performance even more remarkable. 
Moreover, this also translates to the Evaluation Return shown in \autoref{fig:mazes11:test}, where a near optimal performance of around 20 is reached. 
Generally, due to their increased complexity, larger mazes have shown to be less prune to overfitting, especially when training with a pool of random configurations. 
However, still, CROP trained policies, especially using the proposed Radius method show the fastest training convergence and the best generalizing capabilities to unseen configurations.

\section{Conclusion}
Overall, we formalized a method for Compact Reshaped Observation Processing (CROP) and proposed three concrete CROPs applicable to fully observable discrete environments: 
Radius CROP, compressing the observation w.r.t. close positional proximity, Action CROP, compressing the observation w.r.t. interactability, and Object CROP, compressing the observation w.r.t. relations to surrounding objects.
Furthermore, we showed the improvement using any of the proposed CROP over the full observation regarding both the training speed (steps until an optimal policy is found) and, more importantly, the robustness to a distributional shift in the environment in two holey safety environments. 
Finally, benchmark comparisons to the state-of-the-art using augmented data in procedural generated mazes further confirmed the advantages of CROPed observations, showing improved generalization to unseen maze configurations.
Overall, Radius CROP has shown most beneficial, outperforming the full observation in all tested configurations. 

Overall, we believe that CROPing observations to improve their information relevance is a promising approach for improving both the robustness and reliability of reinforcement learning algorithms.

Future work should therefore consider methods and applicability to further observation spaces. 
Also, further CROP methods could be developed with attention to different important features in the observation space. 
Furthermore, the proposed concepts could be applied for training a CROP mechanism for automatic state reduction. 
For example, they could serve as an inspiration for a target of the latent space of an autoencoder, that could be applied to CROP continuous or even partially observable observation spaces. 
Generally, various meta learning concepts may be applied to extend CROP for a more universal applicability. 

\bibliography{CROP.bbl}

\begin{thebibliography}{}

\bibitem[\protect\citeauthoryear{Agarwal \bgroup \em et al.\egroup
  }{2021}]{agarwal2021contrastive}
Rishabh Agarwal, Marlos~C Machado, Pablo~Samuel Castro, and Marc~G Bellemare.
\newblock Contrastive behavioral similarity embeddings for generalization in
  reinforcement learning.
\newblock {\em arXiv preprint arXiv:2101.05265}, 2021.

\bibitem[\protect\citeauthoryear{Amodei \bgroup \em et al.\egroup
  }{2016}]{amodei2016concrete}
Dario Amodei, Chris Olah, Jacob Steinhardt, Paul Christiano, John Schulman, and
  Dan Man{\'e}.
\newblock Concrete problems in ai safety.
\newblock {\em arXiv preprint arXiv:1606.06565}, 2016.

\bibitem[\protect\citeauthoryear{Bartlett \bgroup \em et al.\egroup
  }{2020}]{bartlett2020benign}
Peter~L Bartlett, Philip~M Long, G{\'a}bor Lugosi, and Alexander Tsigler.
\newblock Benign overfitting in linear regression.
\newblock {\em Proceedings of the National Academy of Sciences},
  117(48):30063--30070, 2020.

\bibitem[\protect\citeauthoryear{Bishop and
  Nasrabadi}{2006}]{bishop2006pattern}
Christopher~M Bishop and Nasser~M Nasrabadi.
\newblock {\em Pattern recognition and machine learning}, volume~4.
\newblock Springer, 2006.

\bibitem[\protect\citeauthoryear{Brockman \bgroup \em et al.\egroup
  }{2016}]{gym}
Greg Brockman, Vicki Cheung, Ludwig Pettersson, Jonas Schneider, John Schulman,
  Jie Tang, and Wojciech Zaremba.
\newblock Openai gym, 2016.

\bibitem[\protect\citeauthoryear{Chen \bgroup \em et al.\egroup
  }{2014}]{chen2014cross}
Bor-Chun Chen, Chu-Song Chen, and Winston~H Hsu.
\newblock Cross-age reference coding for age-invariant face recognition and
  retrieval.
\newblock In {\em European conference on computer vision}, pages 768--783.
  Springer, 2014.

\bibitem[\protect\citeauthoryear{Chen \bgroup \em et al.\egroup
  }{2020}]{chen2020gridmask}
Pengguang Chen, Shu Liu, Hengshuang Zhao, and Jiaya Jia.
\newblock Gridmask data augmentation.
\newblock {\em arXiv preprint arXiv:2001.04086}, 2020.

\bibitem[\protect\citeauthoryear{Choi \bgroup \em et al.\egroup
  }{2019}]{choi2019deep}
Jinyoung Choi, Kyungsik Park, Minsu Kim, and Sangok Seok.
\newblock Deep reinforcement learning of navigation in a complex and crowded
  environment with a limited field of view.
\newblock In {\em 2019 International Conference on Robotics and Automation
  (ICRA)}, pages 5993--6000. IEEE, 2019.

\bibitem[\protect\citeauthoryear{Cobbe \bgroup \em et al.\egroup
  }{2020}]{cobbe2020leveraging}
Karl Cobbe, Chris Hesse, Jacob Hilton, and John Schulman.
\newblock Leveraging procedural generation to benchmark reinforcement learning.
\newblock In {\em International conference on machine learning}, pages
  2048--2056. PMLR, 2020.

\bibitem[\protect\citeauthoryear{Ghosh \bgroup \em et al.\egroup
  }{2021}]{ghosh2021generalization}
Dibya Ghosh, Jad Rahme, Aviral Kumar, Amy Zhang, Ryan~P Adams, and Sergey
  Levine.
\newblock Why generalization in rl is difficult: Epistemic pomdps and implicit
  partial observability.
\newblock {\em Advances in Neural Information Processing Systems},
  34:25502--25515, 2021.

\bibitem[\protect\citeauthoryear{Gissl{\'e}n \bgroup \em et al.\egroup
  }{2021}]{gisslen2021adversarial}
Linus Gissl{\'e}n, Andy Eakins, Camilo Gordillo, Joakim Bergdahl, and Konrad
  Tollmar.
\newblock Adversarial reinforcement learning for procedural content generation.
\newblock In {\em 2021 IEEE Conference on Games (CoG)}, pages 1--8. IEEE, 2021.

\bibitem[\protect\citeauthoryear{Hendrycks \bgroup \em et al.\egroup
  }{2021}]{hendrycks2021many}
Dan Hendrycks, Steven Basart, Norman Mu, Saurav Kadavath, Frank Wang, Evan
  Dorundo, Rahul Desai, Tyler Zhu, Samyak Parajuli, Mike Guo, et~al.
\newblock The many faces of robustness: A critical analysis of
  out-of-distribution generalization.
\newblock In {\em Proceedings of the IEEE/CVF International Conference on
  Computer Vision}, pages 8340--8349, 2021.

\bibitem[\protect\citeauthoryear{Jaakkola \bgroup \em et al.\egroup
  }{1994}]{jaakkola1994reinforcement}
Tommi Jaakkola, Satinder Singh, and Michael Jordan.
\newblock Reinforcement learning algorithm for partially observable markov
  decision problems.
\newblock {\em Advances in neural information processing systems}, 7, 1994.

\bibitem[\protect\citeauthoryear{Kaelbling \bgroup \em et al.\egroup
  }{1998}]{kaelbling1998planning}
Leslie~Pack Kaelbling, Michael~L Littman, and Anthony~R Cassandra.
\newblock Planning and acting in partially observable stochastic domains.
\newblock {\em Artificial intelligence}, 101(1-2):99--134, 1998.

\bibitem[\protect\citeauthoryear{Karystinos and
  Pados}{2000}]{karystinos2000overfitting}
George~N Karystinos and Dimitrios~A Pados.
\newblock On overfitting, generalization, and randomly expanded training sets.
\newblock {\em IEEE Transactions on Neural Networks}, 11(5):1050--1057, 2000.

\bibitem[\protect\citeauthoryear{Kostrikov \bgroup \em et al.\egroup
  }{2020}]{kostrikov2020image}
Ilya Kostrikov, Denis Yarats, and Rob Fergus.
\newblock Image augmentation is all you need: Regularizing deep reinforcement
  learning from pixels.
\newblock {\em arXiv preprint arXiv:2004.13649}, 2020.

\bibitem[\protect\citeauthoryear{Laskin \bgroup \em et al.\egroup
  }{2020}]{laskin2020reinforcement}
Misha Laskin, Kimin Lee, Adam Stooke, Lerrel Pinto, Pieter Abbeel, and Aravind
  Srinivas.
\newblock Reinforcement learning with augmented data.
\newblock {\em Advances in neural information processing systems},
  33:19884--19895, 2020.

\bibitem[\protect\citeauthoryear{Leike \bgroup \em et al.\egroup
  }{2017}]{leike2017ai}
Jan Leike, Miljan Martic, Victoria Krakovna, Pedro~A Ortega, Tom Everitt,
  Andrew Lefrancq, Laurent Orseau, and Shane Legg.
\newblock Ai safety gridworlds.
\newblock {\em arXiv preprint arXiv:1711.09883}, 2017.

\bibitem[\protect\citeauthoryear{Malinin \bgroup \em et al.\egroup
  }{2021}]{malinin2021shifts}
Andrey Malinin, Neil Band, German Chesnokov, Yarin Gal, Mark~JF Gales, Alexey
  Noskov, Andrey Ploskonosov, Liudmila Prokhorenkova, Ivan Provilkov, Vatsal
  Raina, et~al.
\newblock Shifts: A dataset of real distributional shift across multiple
  large-scale tasks.
\newblock {\em arXiv preprint arXiv:2107.07455}, 2021.

\bibitem[\protect\citeauthoryear{Mazoure \bgroup \em et al.\egroup
  }{2021}]{mazoure2021cross}
Bogdan Mazoure, Ahmed~M Ahmed, Patrick MacAlpine, R~Devon Hjelm, and Andrey
  Kolobov.
\newblock Cross-trajectory representation learning for zero-shot generalization
  in rl.
\newblock {\em arXiv preprint arXiv:2106.02193}, 2021.

\bibitem[\protect\citeauthoryear{Mnih \bgroup \em et al.\egroup
  }{2016}]{mnih2016asynchronous}
Volodymyr Mnih, Adria~Puigdomenech Badia, Mehdi Mirza, Alex Graves, Timothy
  Lillicrap, Tim Harley, David Silver, and Koray Kavukcuoglu.
\newblock {Asynchronous} {Methods} for {Deep} {Reinforcement} {Learning}.
\newblock In {\em International Conference on Machine Learning}, 2016.

\bibitem[\protect\citeauthoryear{Najar and
  Chetouani}{2021}]{najar2021reinforcement}
Anis Najar and Mohamed Chetouani.
\newblock Reinforcement learning with human advice: a survey.
\newblock {\em Frontiers in Robotics and AI}, 8:584075, 2021.

\bibitem[\protect\citeauthoryear{Pimentel \bgroup \em et al.\egroup
  }{2014}]{pimentel2014review}
Marco~AF Pimentel, David~A Clifton, Lei Clifton, and Lionel Tarassenko.
\newblock A review of novelty detection.
\newblock {\em Signal processing}, 99:215--249, 2014.

\bibitem[\protect\citeauthoryear{Pinto \bgroup \em et al.\egroup
  }{2017}]{pinto2017robust}
Lerrel Pinto, James Davidson, Rahul Sukthankar, and Abhinav Gupta.
\newblock Robust adversarial reinforcement learning.
\newblock In {\em International Conference on Machine Learning}, pages
  2817--2826. PMLR, 2017.

\bibitem[\protect\citeauthoryear{Puterman}{1990}]{puterman1990markov}
Martin~L Puterman.
\newblock Markov decision processes.
\newblock {\em Handbooks in operations research and management science},
  2:331--434, 1990.

\bibitem[\protect\citeauthoryear{Quinonero-Candela \bgroup \em et al.\egroup
  }{2008}]{quinonero2008dataset}
Joaquin Quinonero-Candela, Masashi Sugiyama, Anton Schwaighofer, and Neil~D
  Lawrence.
\newblock {\em Dataset shift in machine learning}.
\newblock Mit Press, 2008.

\bibitem[\protect\citeauthoryear{Raffin \bgroup \em et al.\egroup
  }{2021}]{stable-baselines3}
Antonin Raffin, Ashley Hill, Adam Gleave, Anssi Kanervisto, Maximilian
  Ernestus, and Noah Dormann.
\newblock Stable-baselines3: Reliable reinforcement learning implementations.
\newblock {\em Journal of Machine Learning Research}, 22(268):1--8, 2021.

\bibitem[\protect\citeauthoryear{Raileanu \bgroup \em et al.\egroup
  }{2020}]{raileanu2020automatic}
Roberta Raileanu, Max Goldstein, Denis Yarats, Ilya Kostrikov, and Rob Fergus.
\newblock Automatic data augmentation for generalization in deep reinforcement
  learning.
\newblock {\em arXiv preprint arXiv:2006.12862}, 2020.

\bibitem[\protect\citeauthoryear{Ramanan \bgroup \em et al.\egroup
  }{2021}]{ramanan2021real}
Nandini Ramanan, Rasool Tahmasbi, Marjorie Sayer, Deokwoo Jung, Shalini
  Hemachandran, and Claudionor~Nunes Coelho~Jr.
\newblock Real-time drift detection on time-series data.
\newblock {\em arXiv preprint arXiv:2110.06383}, 2021.

\bibitem[\protect\citeauthoryear{Raskutti \bgroup \em et al.\egroup
  }{2014}]{raskutti2014early}
Garvesh Raskutti, Martin~J Wainwright, and Bin Yu.
\newblock Early stopping and non-parametric regression: an optimal
  data-dependent stopping rule.
\newblock {\em The Journal of Machine Learning Research}, 15(1):335--366, 2014.

\bibitem[\protect\citeauthoryear{Schulman \bgroup \em et al.\egroup
  }{2017}]{schulman2017proximal}
John Schulman, Filip Wolski, Prafulla Dhariwal, Alec Radford, and Oleg Klimov.
\newblock Proximal policy optimization algorithms.
\newblock {\em arXiv preprint arXiv:1707.06347}, 2017.

\bibitem[\protect\citeauthoryear{Shorten and
  Khoshgoftaar}{2019}]{shorten2019survey}
Connor Shorten and Taghi~M Khoshgoftaar.
\newblock A survey on image data augmentation for deep learning.
\newblock {\em Journal of big data}, 6(1):1--48, 2019.

\bibitem[\protect\citeauthoryear{Spaan}{2012}]{spaan2012partially}
Matthijs~TJ Spaan.
\newblock Partially observable markov decision processes.
\newblock In {\em Reinforcement Learning}, pages 387--414. Springer, 2012.

\bibitem[\protect\citeauthoryear{Srivastava \bgroup \em et al.\egroup
  }{2014}]{srivastava2014dropout}
Nitish Srivastava, Geoffrey Hinton, Alex Krizhevsky, Ilya Sutskever, and Ruslan
  Salakhutdinov.
\newblock Dropout: a simple way to prevent neural networks from overfitting.
\newblock {\em The journal of machine learning research}, 15(1):1929--1958,
  2014.

\bibitem[\protect\citeauthoryear{Sutton and
  Barto}{2018}]{sutton2018reinforcement}
Richard~S Sutton and Andrew~G Barto.
\newblock {\em Reinforcement learning: An introduction}.
\newblock MIT press, 2018.

\bibitem[\protect\citeauthoryear{Sutton \bgroup \em et al.\egroup
  }{2000}]{sutton2000policy}
Richard~S Sutton, David~A McAllester, Satinder~P Singh, and Yishay Mansour.
\newblock {Policy} {Gradient} {Methods} for {Reinforcement} {Learning} with
  {Function} {Approximation}.
\newblock In S.~Solla, T.~Leen, and K.~M\"{u}ller, editors, {\em Advances in
  Neural Information Processing Systems}, volume~12, pages 1057--1063. MIT
  Press, 2000.

\bibitem[\protect\citeauthoryear{Thulasidasan \bgroup \em et al.\egroup
  }{2021}]{thulasidasan2021effective}
Sunil Thulasidasan, Sushil Thapa, Sayera Dhaubhadel, Gopinath Chennupati,
  Tanmoy Bhattacharya, and Jeff Bilmes.
\newblock An effective baseline for robustness to distributional shift.
\newblock In {\em 2021 20th IEEE International Conference on Machine Learning
  and Applications (ICMLA)}, pages 278--285. IEEE, 2021.

\bibitem[\protect\citeauthoryear{Tobin \bgroup \em et al.\egroup
  }{2017}]{tobin2017domain}
Josh Tobin, Rachel Fong, Alex Ray, Jonas Schneider, Wojciech Zaremba, and
  Pieter Abbeel.
\newblock Domain randomization for transferring deep neural networks from
  simulation to the real world.
\newblock In {\em 2017 IEEE/RSJ international conference on intelligent robots
  and systems (IROS)}, pages 23--30. IEEE, 2017.

\bibitem[\protect\citeauthoryear{Vlassis \bgroup \em et al.\egroup
  }{2012}]{vlassis2012computational}
Nikos Vlassis, Michael~L Littman, and David Barber.
\newblock On the computational complexity of stochastic controller optimization
  in pomdps.
\newblock {\em ACM Transactions on Computation Theory (TOCT)}, 4(4):1--8, 2012.

\bibitem[\protect\citeauthoryear{Wang \bgroup \em et al.\egroup
  }{2020}]{wang2020generalizing}
Yaqing Wang, Quanming Yao, James~T Kwok, and Lionel~M Ni.
\newblock Generalizing from a few examples: A survey on few-shot learning.
\newblock {\em ACM computing surveys (csur)}, 53(3):1--34, 2020.

\bibitem[\protect\citeauthoryear{Yarats \bgroup \em et al.\egroup
  }{2021}]{yarats2021mastering}
Denis Yarats, Rob Fergus, Alessandro Lazaric, and Lerrel Pinto.
\newblock Mastering visual continuous control: Improved data-augmented
  reinforcement learning.
\newblock {\em arXiv preprint arXiv:2107.09645}, 2021.

\bibitem[\protect\citeauthoryear{Ying}{2019}]{ying2019overview}
Xue Ying.
\newblock An overview of overfitting and its solutions.
\newblock {\em Journal of Physics: Conference Series}, 1168(2):022022, 2019.

\bibitem[\protect\citeauthoryear{Zhang \bgroup \em et al.\egroup
  }{2020}]{zhang2020learning}
Amy Zhang, Rowan McAllister, Roberto Calandra, Yarin Gal, and Sergey Levine.
\newblock Learning invariant representations for reinforcement learning without
  reconstruction.
\newblock {\em arXiv preprint arXiv:2006.10742}, 2020.

\bibitem[\protect\citeauthoryear{Zhao \bgroup \em et al.\egroup
  }{2020}]{zhao2020sim}
Wenshuai Zhao, Jorge~Pe{\~n}a Queralta, and Tomi Westerlund.
\newblock Sim-to-real transfer in deep reinforcement learning for robotics: a
  survey.
\newblock In {\em 2020 IEEE Symposium Series on Computational Intelligence
  (SSCI)}, pages 737--744. IEEE, 2020.

\end{thebibliography}
\end{document}